\PassOptionsToPackage{table}{xcolor}

\documentclass[10pt,twocolumn,letterpaper]{article}

\usepackage[pagenumbers]{cvpr} 

\usepackage{multirow}
\usepackage{makecell}

\usepackage{bbding}
\usepackage{amssymb}
\usepackage{pifont}
\newcommand{\cmark}{\ding{51}}%
\newcommand{\xmark}{\ding{55}}%
\usepackage[table]{xcolor}
\usepackage{xcolor}
\usepackage{array}

\usepackage[accsupp]{axessibility} 

\definecolor{cvprblue}{rgb}{0.21,0.49,0.74}
\usepackage[pagebackref,breaklinks,colorlinks,citecolor=cvprblue]{hyperref}

\def\paperID{12220} 
\def\confName{CVPR}
\def\confYear{2024}
\def\confDataName{PKU-DyMVHumans}

\title{\confDataName: A Multi-View Video Benchmark for\\ High-Fidelity Dynamic Human Modeling}

\author{Xiaoyun Zheng $^{1, 2}$  \quad Liwei Liao $^{1, 2}$ \quad Xufeng Li $^{3}$ \quad Jianbo Jiao $^{4}$   \\
        Rongjie Wang $^{2}$ \quad Feng Gao $^{5}$  \quad  Shiqi Wang $^{3}$ \quad Ronggang Wang $^{1 \dagger}$ \\      
$^{1}$ Peking University Shenzhen Graduate School \quad $^{2}$ Peng Cheng Laboratory \\
$^{3}$ City University of Hong Kong \quad $^{4}$ University of Birmingham \quad $^{5}$ Peking University   \\
{\tt\small xyun\_z@stu.pku.edu.cn} \quad {\tt\small \{levio, gaof\}@pku.edu.cn} \quad  {\tt\small xufengli2-c@my.cityu.edu.hk} \\
{\tt\small j.jiao@bham.ac.uk}  \quad {\tt\small wangrj@pcl.ac.cn} \quad {\tt\small shiqwang@cityu.edu.hk} \quad {\tt\small rgwang@pkusz.edu.cn} 
}

\begin{document}

\twocolumn[{%
\renewcommand\twocolumn[1][]{#1}%
\maketitle
\begin{center}
\vspace{-12mm}
  \centering
      \captionsetup{type=figure}
   \includegraphics[width=\linewidth]{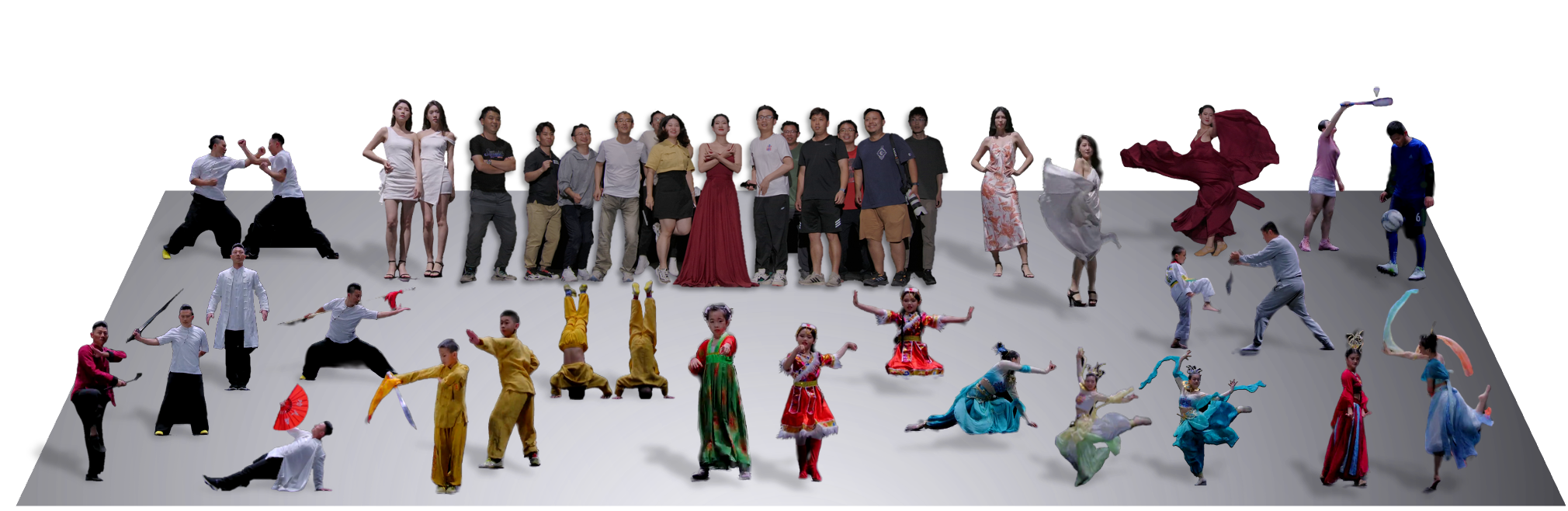} 
   \vspace{-6mm}
   \captionof{figure}{We present {\bf \confDataName}, a versatile human-centric dataset designed for high-fidelity reconstruction and rendering of dynamic human performances from dense multi-view videos. It comprises 32 humans across 45 different dynamic scenarios, each featuring highly detailed appearances and complex human motions. }
   \label{fig:snapshot}
\end{center}%
}]

\begin{abstract}
High-quality human reconstruction and photo-realistic rendering of a dynamic scene is a long-standing problem in computer vision and graphics. Despite considerable efforts invested in developing various capture systems and reconstruction algorithms, recent advancements still struggle with loose or oversized clothing and overly complex poses. In part, this is due to the challenges of acquiring high-quality human datasets. To facilitate the development of these fields, in this paper, we present \confDataName, a versatile human-centric dataset for high-fidelity reconstruction and rendering of dynamic human scenarios from dense multi-view videos. It comprises 8.2 million frames captured by more than 56 synchronized cameras across diverse scenarios. These sequences comprise 32 human subjects across 45 different scenarios, each with a high-detailed appearance and realistic human motion. Inspired by recent advancements in neural radiance field (NeRF)-based scene representations, we carefully set up an off-the-shelf framework that is easy to provide those state-of-the-art NeRF-based implementations and benchmark on \confDataName~dataset. It is paving the way for various applications like fine-grained foreground/background decomposition, high-quality human reconstruction and photo-realistic novel view synthesis of a dynamic scene. Extensive studies are performed on the benchmark, demonstrating new observations and challenges that emerge from using such high-fidelity dynamic data. The project page and data is available at: \url{https://pku-dymvhumans.github.io}.
\end{abstract}

\vspace{-5mm}
\section{Introduction}
\label{sec:intro}
We are entering an era in which the distinction between virtual and real worlds is becoming increasingly blurred. High-quality reconstruction and photo-realistic rendering of human activities are crucial for many immersive applications, including AR/VR, 3D content production, and entertainment. However, the process of reconstructing human activities and providing photo-realistic rendering from multiple viewpoints is currently a cutting-edge but challenging technique. The limited availability of real-world datasets is impeding progress in this critical task.

Early solutions adopted multi-view stereo techniques to explicitly reconstruct textured meshes\cite{temporally, high_fvv, 2019relightables}, or employed view interpolation for image-based rendering\cite{lumigraph, editable}. However, these methods require pre-scanned templates or dense camera rigs (up to 100 cameras \cite{2019relightables}), which are expensive and not easily portable or affordable. To simplify the capture systems, some works have used commodity depth sensors to build real-time reconstruction systems \cite{fusion4d, function4d}, but they are limited by inherent self-occlusion constraints. Moreover, these approaches are prone to producing reconstruction errors and rendering artifacts in scenes with thin structures, specular surfaces, and topological changes.

The recent advancements in learning-based techniques have made it possible to achieve robust human attribute reconstruction \cite{pifu, pifuhd, monoport} using only RGB input. Specifically, the methods PIFu \cite{pifu} and PIFuHD \cite{pifuhd} utilize pixel-aligned implicit functions to accurately reconstruct clothed humans with intricate geometric details. However, these approaches struggle with generating realistic appearances, largely due to their reliance on implicit texture representation. On the other hand, recent advancements in neural rendering techniques have showcased impressive capabilities in synthesizing novel views of general static scenes using neural radiance field (NeRF) representations \cite{nerf}. This approach has also shown promising results in human modeling \cite{neurabody, humannerf, enerf}, although it still relies on relatively dense multiview videos as input. To address this limitation, some recent studies \cite{tava, animatable} propose using human body priors to assist in learning human representations.

Methods that utilize radiance fields aim to optimize both the input and output representation. However, these problems still remain open. Currently, it is still challenging to achieve a balance between realism, robustness to complex poses and clothing, reasonable computation time, and compatibility with complex scenes. One of the challenges stems from the flexibility of humans, as they move in complex ways against natural backgrounds, and their clothing and muscles deform. Additionally, other factors such as occlusion may require comprehensive scene modeling beyond just the humans present. Another pressing issue is the lack of high-quality, high-detail datasets of complex scenarios, as well as the need for accurate human pose and segmentation to training networks, which makes it difficult to accurately evaluate multi-person performance capture systems. Looking ahead, we anticipate the development of a new versatile model that is powerful yet general in terms of both appearance rendering and motion modeling. This model should be capable of generating realistic dynamic details without the need for pre-scanning or pre-processing efforts.

In this work, we propose \confDataName, a versatile human-centric dataset that includes 32 dynamic humans performing various actions and wearing different clothing styles, as shown in \cref{fig:snapshot}. The dataset consists of approximately 8.2 million frames and aims to address the lack of large-scale and high-fidelity human performance datasets. Compared to existing benchmark datasets, \confDataName~dataset offers appealing characteristics: 
\textbf{1) High-fidelity human performance}: We construct a professional multi-view system to capture humans in motion, which contains 56/60 synchronous cameras with 1080P or 4K resolution.
\textbf{2) High-detailed appearance}: It captures complex cloth deformation, and intricate texture details, such as delicate satin ribbon and vivid textures of classical headwear. 
\textbf{3) Complex human motion}: It covers a wide range of special costume performances, artistic movements, and sports activities. 
\textbf{4) Human-object/scene interactions}: These include human-object interactions, as well as challenging multi-person interactions and complex scene effects (\eg, lighting, shadows, and smoking). 

\begin{figure}
  \centering
   \includegraphics[width=1.02\linewidth]{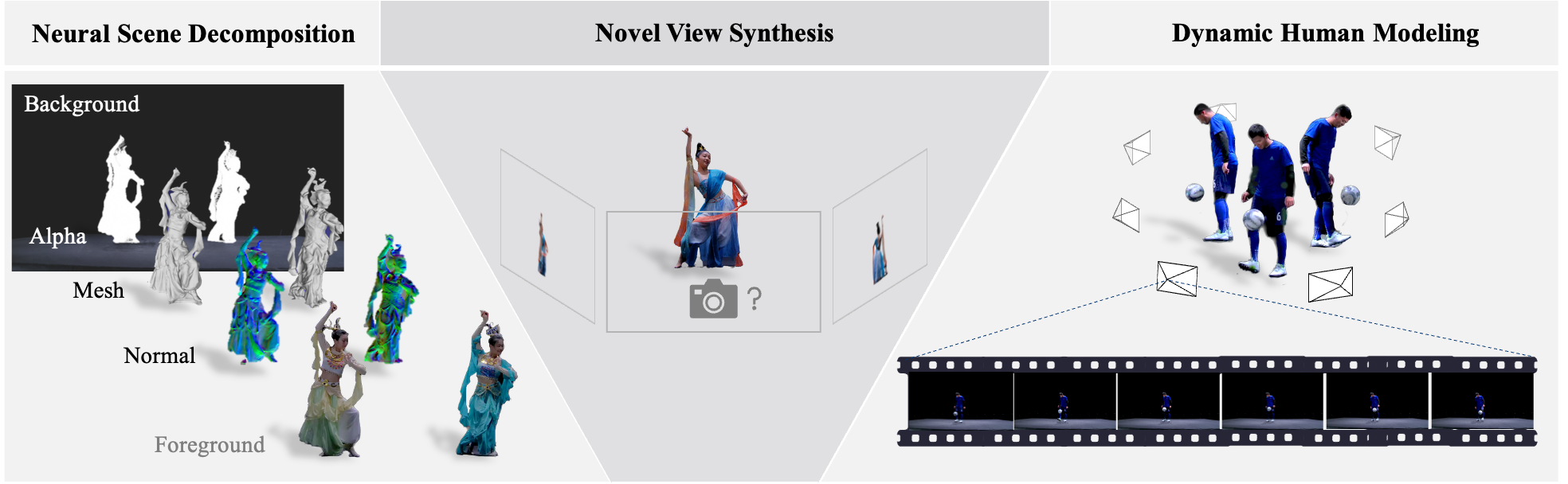}
   \vspace{-6mm}
   \caption{{\bf Research with \confDataName.} It supports various research topics, including neural scene decomposition, novel view synthesis, and dynamic human modeling.}
   \label{fig:usage}
   \vspace{-4mm}
\end{figure}

The primary goal of \confDataName~is to enable high-fidelity reconstruction and rendering of human performances from dense multi-view videos. To leverage the extensive exploration space offered by our dataset, we develop a unified framework that enables the implementation and evaluation of state-of-the-art NeRF-based approaches on \confDataName. As depicted in \cref{fig:usage}, the framework opens up possibilities for various applications, including fine-grained foreground/background decomposition, high-quality human reconstruction, and photo-realistic synthesis of dynamic humans. Extensive studies have been conducted on the benchmark, revealing new insights and challenges associated with the use of such high-fidelity dynamic data.

\section{Related Work}
\label{sec:related}

\begin{table*}[h]
\scriptsize 
\setlength{\tabcolsep}{5pt} 
\setlength{\extrarowheight}{0.2pt} 
\centering
\caption{\textbf{Comparisons of multi-view human datasets.} We compare the proposed dataset with previous human-centric multi-view datasets in terms of scale, attribute, and resolution. The {\colorbox {gray!20}{gray color}} indicates the marker-based capture data with reliable human motion or pre-scanned humans reconstructed from depth sensors or stereo camera arrays. 
Differing from these efforts, \confDataName~features high-fidelity dynamic sequences with high-detailed appearance, complex human motions, as well as challenging multi-person interactions. } 
\vspace{-2mm}
\label{tab:overview}
\begin{tabular}{llllllccccll}
\toprule
\textbf{Dataset}  &\textbf{\#Cam.} &\textbf{\#Subj.}  &\textbf{\#Scene} & \textbf{\#Seq.} & \textbf{\#Frame}  & \textbf{Dynamic} & \textbf{Interaction} & \textbf{High-detail} &\textbf{Multi-person} & \textbf{Resolution}  \\ 
\midrule
\rowcolor{gray!20}Human3.6M\cite{h36m}       & 4   & 11     & 15        & 839             & 3.6M    & \cmark & \cmark & \xmark & \xmark   & 1000×1002  \\ 
\rowcolor{gray!20}THUman2.0\cite{thuman2.0}  & 60  & 200    & -         & -               & -       & \xmark & \xmark & \xmark & \xmark   & 512×512    \\ 
\rowcolor{gray!20}Hi4D\cite{hi4d}            & 8   & 40     & 20        & 100             & 11K     & \cmark & \cmark & \xmark & \cmark   & 940×1280   \\ 
\rowcolor{gray!20}DNA-Rendering\cite{dna}    & 60  & 500    & 1,187     & 25,320          & 67.5M   & \cmark & \cmark & \cmark & \xmark   & 4096×3000, 2448×2048 \\
\rowcolor{gray!20}DynaCap\cite{dynacap}   & 50-101  & 4     & 5         & 440             & 11.8M   & \cmark & \xmark & \xmark & \xmark   & 1285×940   \\
\rowcolor{gray!20}MultiHuman\cite{multihuman} & 128 & 50    & -         & -               & -       & \xmark & \cmark & \xmark & \cmark   &   -        \\ 
\rowcolor{gray!20}THuman5.0\cite{thuman5.0}  & 32   & 10    & 10        & 320             & 96K     & \cmark & \xmark & \cmark & \xmark   & 4096×3000  \\
\rowcolor{gray!20}HuMMan\cite{HuMMan}        & 11   & 132   & 20        & 4,466           & 278K    & \cmark & \cmark & \xmark & \xmark   & 1920×1080  \\
\rowcolor{gray!20}UltraStage\cite{relightable} & 32 & 100   & 20        & -               & 192K    & \xmark & \xmark & \cmark & \xmark   & 7860×4320  \\
\rowcolor{gray!20}Actors-HQ\cite{humanrf}    & 160  & 8     & 21        & 2,560           & 39.8K   & \cmark & \xmark & \cmark & \xmark   & 4096×3072  \\ 
HUMBI\cite{humbi}                            & 107  & 772   & 4         & -               & 26M     & \xmark & \xmark & \cmark & \xmark   & 1920×1080  \\
NHR\cite{nhr}                               & 56/72 & 4     & 3         & 320             & 47K     & \cmark & \cmark & \xmark & \xmark   & 1024×768   \\
AIST++\cite{aist}                            & 9    & 30    & 10        & 1,408           & 10.1M   & \cmark & \cmark & \xmark & \cmark   & 1920×1080   \\
ZJU\_Mocap\cite{neurabody}                   & 21   & 9     & 6         & \~207           & 185K    & \cmark & \xmark & \xmark & \xmark   & 1024×1024   \\
THuman4.0\cite{thuman4.0}                    & 24   & 3     & 3         & 320             & 250K    & \cmark & \xmark & \xmark & \xmark   & 1330×1150   \\
ENeRF-Outdoor\cite{enerf}                    & 18   & 7     & 4         & 72              & 86K     & \cmark & \cmark & \xmark & \cmark   & 1920×1080    \\
FreeMan\cite{freeman}                        & 8    & 40    & 123       & 8,000           & 11.3M   & \cmark & \cmark & \xmark & \xmark   & 1920×1080    \\
\midrule
\textbf{\confDataName~(Ours)}              & 56/60  & 32    & 45        & 2,668            & 8.2M    & \cmark & \cmark & \cmark & \cmark  & 3840×2160,  1920×1080  \\
\bottomrule
\end{tabular}

\vspace{-4mm}
\end{table*}

{\bf Human performance capture and datasets.} Human performance capture, which involves capturing the full pose and non-rigid surface deformation of people wearing regular clothing in a space-time coherent 4D manner, has revolutionized the film and gaming industry in recent years \cite{deepcap}. The early existing methods learn from high-quality data obtained from optical marker-based capture systems~\cite{h36m, thuman2.0}. Other recent methods utilize depth sensor~\cite{thuman5.0, hi4d, dna} or volumetric data~\cite{deepcap, dynacap} to achieve reliable human motion and geometry. However, the sophisticated setup restricts their practical deployment. With the emergence of neural rendering techniques, rendering realistic humans directly from images has become a trend. Such a setting~\cite{markerless} usually requires the dataset equipped with high-quality dense view images~\cite{humbi, nhr} or accurate annotations like human body keypoints and foreground segmentation~\cite{aist, neurabody, thuman4.0, freeman}. However, recent methods still often struggle to reconstruct models accurately when the clothes are loose, oversized, or when the poses are too complex. Differing from these efforts, we aim to provide a versatile human-centric dataset designed for high-fidelity reconstruction and rendering of dynamic human performances from dense multi-view videos. The unfold comparisons between \confDataName~and the existing datasets are given in~\cref{tab:overview}. When compared to the published benchmark datasets, \confDataName~features high-detailed appearance, complex motion, as well as challenging human-object interactions, multi-person interactions and complex scene effects (\eg, lighting, shadows, and smoking).

{\noindent\bf Neural implicit representations.} In the domain of photo-realistic novel view synthesis and 3D scene modeling, differentiable neural rendering based on various data proxies has achieved impressive results and gained popularity. Traditional methods rely on explicit geometric representations, such as depth maps \cite{rethinking}, point cloud \cite{nhr}, meshes \cite{learning_mesh}, and voxel grids \cite{voxel}. Recently, coordinate-based networks have become a popular choice for implicit 3D scene representations, such as radiance fields \cite{nerf}, signed distance fields (SDF) \cite{deepsdf, gens}, or occupancy \cite{occupancy}. The emerging neural implicit representation approaches show promising results in novel view synthesis \cite{nerf, nsvf, instant_ngp} and high-quality 3D reconstruction from multi-view images \cite{sdfdiff, neus, neus2}. In the pioneering work of NeRF \cite{nerf}, a Multi-Layer Perceptron (MLP) is trained to encode a radiance field reconstructed from a set of input RGB images. However, it cannot extract high-quality surfaces due to the lack of surface constraints in the geometry representation. NeuS \cite{neus} addresses this limitation by representing the 3D surface as an SDF for high-quality geometry reconstruction. However, the explicit integration in NeuS makes it computationally intensive, resulting in slow training and limited applicability to static scene reconstruction. To overcome the slow training of deep coordinate-based MLPs, Instant-NGP \cite{instant_ngp} proposes a multi-resolution hash encoding technique, which has been proven effective in accelerating the training process.

{\noindent\bf Free view synthesis for dynamic scenes.} Photo-realistic rendering of dynamic scenes from a set of input images is necessary for many applications. For dynamic scene modeling, some methods \cite{neural_flow, neural3d, video_nerf} consider dynamic scenes as a 4D domain and add the time dimension to the input spatial coordinate. This approach enables the implementation of space-time radiance fields. In the field of human rendering, several approaches \cite{neuralactor, neurabody, mono_humannerf} utilize human priors to model human motions and achieve free-viewpoint rendering of dynamic human. HumanNeRF \cite{mono_humannerf} demonstrates the ability to render realistic humans from monocular video sequences. While these methods produce impressive rendering results, their training process is time-consuming. To address the need for fast dynamic scene reconstruction, NeuS2 \cite{neus2} integrates multi-resolution hash encoding into SDF and proposes an incremental training strategy with a global transformation prediction component. This approach leverages shared geometry and appearance information in two consecutive frames. Besides, Tensor4D \cite{tensor4d} proposes a hierarchical tri-projection decomposition method to learn high-quality dynamic scenes representation from sparse-view videos. This method models a 4D tensor using nine 2D feature planes, capturing spatio-temporal information in a compact and memory-efficient manner. However, they generally handle videos with short frames.

\begin{figure*}[htbp]
  \centering
   \includegraphics[width=0.98\linewidth]{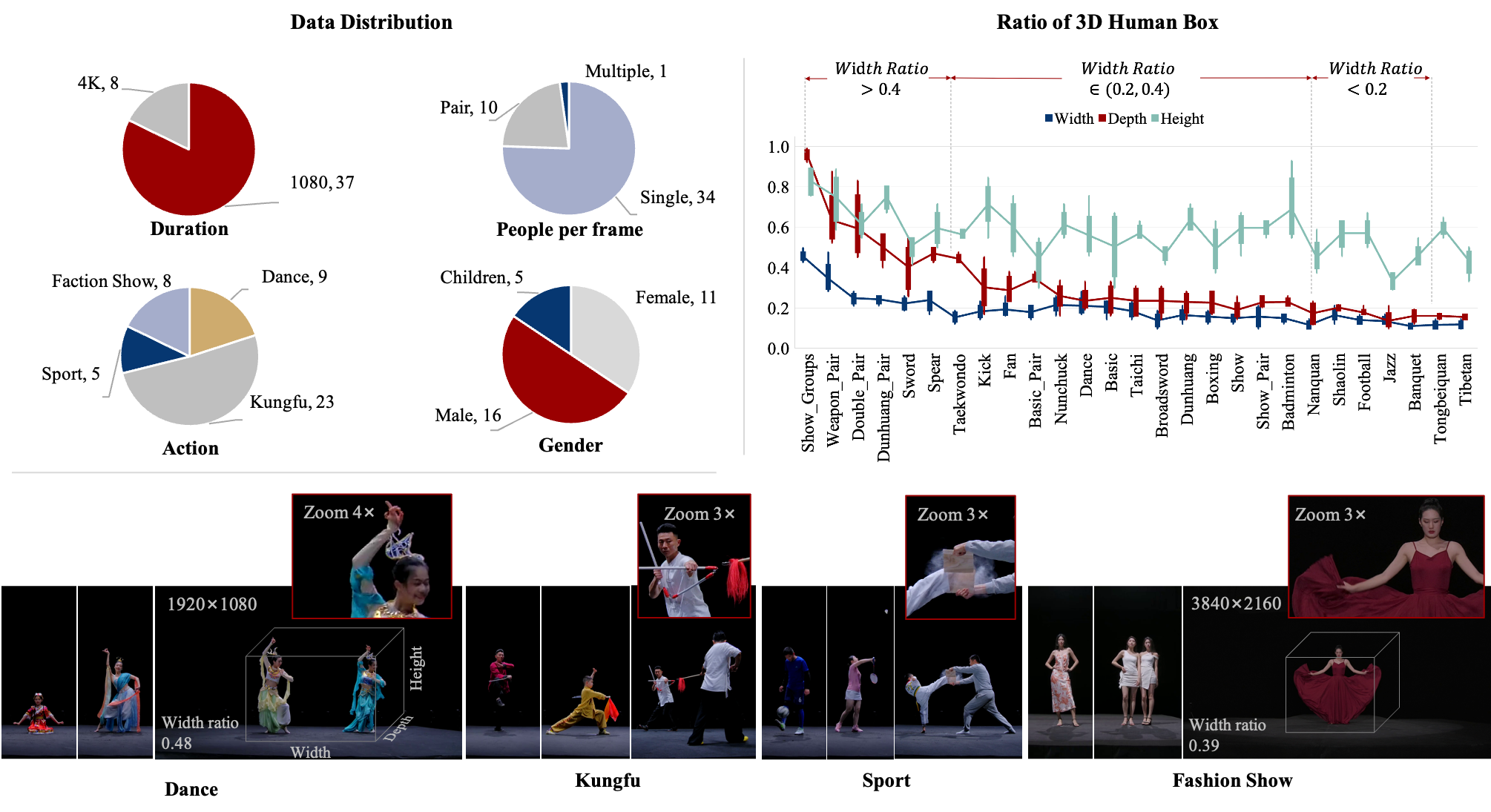}
   
   \vspace{-2mm}
   \caption{Category definition and distribution of the proposed \confDataName~dataset.}
   \label{fig:distribution}
   \vspace{-3mm}
\end{figure*}

\section{The \confDataName~Dataset}
\label{sec:overview}

\subsection{Data Capturing and Processing}
{\bf Data capturing setting.} Our goal is to design a system that can acquire human-centric datasets for high-quality human reconstruction and photo-realistic novel view synthesis in markerless multi-view capture settings. Our capture system is built on an indoor stage, lit by spotlights from above and equipped with a circular camera array. 
(1) For the 1080P sequences category, the camera system comprises 60 Z CAM E2 cameras operating at a resolution of 1920×1080 and 25 FPS. 
The 60 cameras are evenly distributed in a circle around the players, with each camera positioned approximately 6 meters from the system center. 
(2) For the 4K Studio, players stand on the stage are lit by a follow spotlight. 
This setup includes 56 calibrated cameras positioned in a large arc around the players. The distance from each camera to the system center is approximately 6 meters. 
In both situations mentioned above, we capture multiple sequences of challenging human performances, such as dance, kungfu, sport, and fashion shows.

{\noindent\bf Sparse reconstruction.} Following the procedure outlined in \cite{nerf, neus}, we utilize the COLMAP Structure-from-Motion (SfM) algorithm \cite{sfm} to calibrate camera intrinsic and extrinsic parameters. Our goal is to enable a neural processing pipeline capable of accurately reconstructing both static and dynamic scenes from multiple views using state-of-the-art neural implicit representation methods such as NeuS and NeRF-based approaches. Furthermore, we provide a data conversion tool that transforms the sparse model output from SfM into a format compatible with Instant-NGP \cite{instant_ngp}, NeuS, and Tensor4D \cite{tensor4d}.

{\noindent\bf Foreground object segmentation.} Each frame extracted from the original video is passed to BGMv2 \cite{bgmv2} to generate the binary foreground object mask. This not only improves dense reconstruction but also contributes to the subsequent step of dynamic human reconstruction. In this context, our contemporary work Surface-SOS \cite{sos} aims for delicate segmentation by leveraging the cross-view consistency of neural implicit surface representation from a sparse set of posed images (Sec.~\ref{subsec:nsd}). 

\begin{figure*}[htbp]
  \centering
   \includegraphics[width=\linewidth]{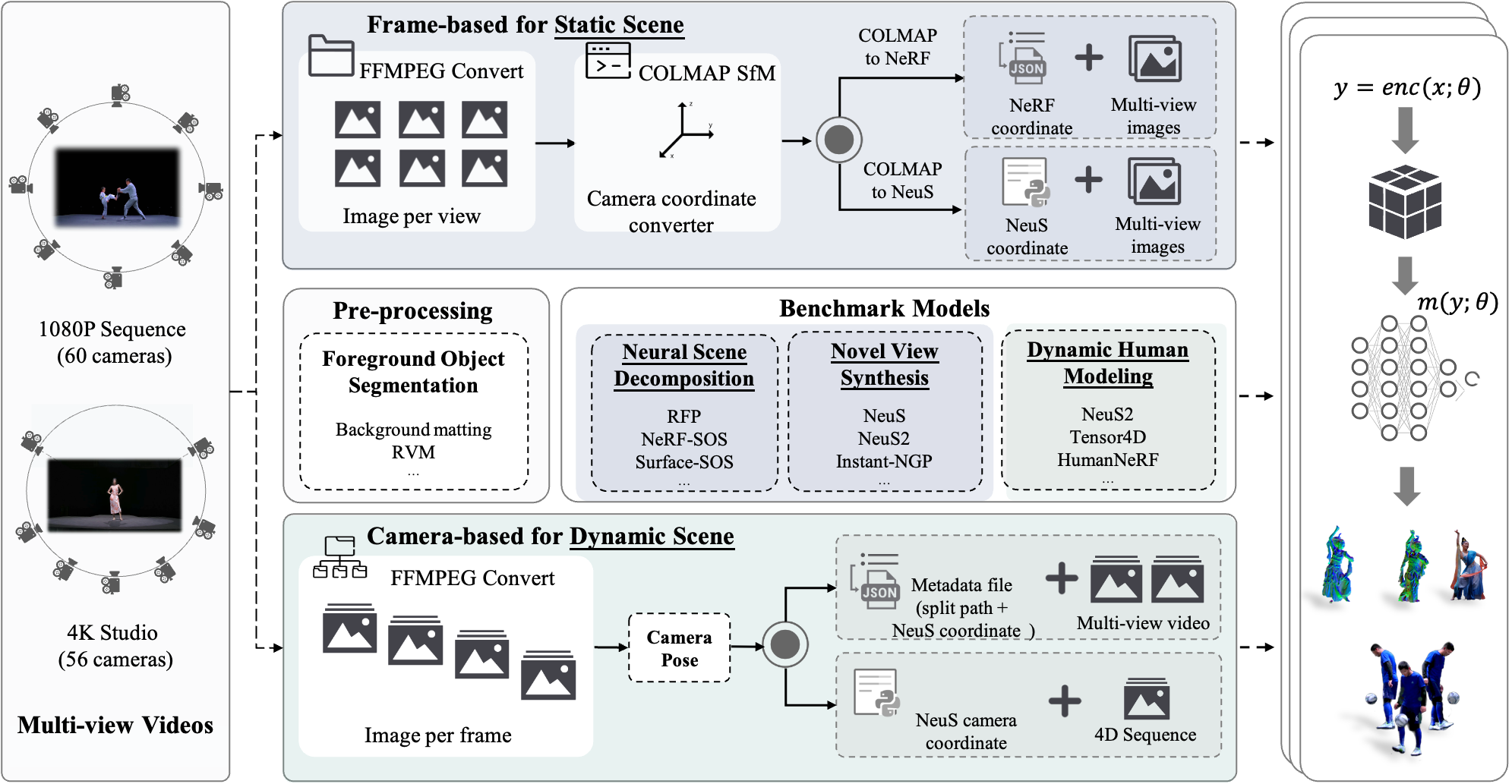}
   
   \vspace{-2mm}
   \caption{{\bf Benchmarks pipeline of \confDataName.} Given a multi-view video input, the first step is to extract the frames and estimate the foreground object mask and camera parameters. Specifically, BGMv2 \cite{bgmv2} is used to generate the binary foreground object mask. Afterwards, COLMAP SfM \cite{sfm} is used to estimate camera parameters and generate a sparse point cloud. Using these components, we have constructed three benchmarks. (a) The implementation of NeRF by Instant-NGP requires providing initial camera parameters in JSON file format compatible with the original NeRF codebase. (b) In addition to RGB and mask images, the NeuS implementation expects a camera file that contains a projection matrix and a normalization matrix for each image. (c) We also provide data conversion from NeuS to NeuS2 and Tensor4D format for specifying dynamic scenes.
   }
   \label{fig:pipeline}
   \vspace{-3mm}
\end{figure*}

\subsection{Dataset Statistics and Distribution}
The \confDataName~dataset consists of 45 different dynamic scenarios, totalling approximately 8.2 million frames of recording. The sequences feature performers in different locations, engaging in various actions and clothing styles. There are 32 professional players, including 16 males, 11 females, and 5 children, performing 4 different action types: dance, kungfu, sport, and fashion show\footnote{All the actors have given consent in signed written forms to the use of their recordings.}. As shown in \cref{fig:distribution}, the diversity of \confDataName~stems from the wide variations in neural human modeling across different motion categories, as well as variances in human size, and character poses. This makes \confDataName~a more challenging dataset compared to previous human datasets, as it requires higher generalization abilities from human-object reconstruction and rendering methods. The average ratios for human width, depth and height are 0.31, 0.19, and 0.57, respectively. For the reason that human bodies are typically long and thin, the depth ratios of humans (3D bounding box depth/image width) tend to be more concentrated in smaller proportions. These ratios indicate that our dataset presents challenges beyond scale, including appearance diversity and pose complexity. 

In the 1080P sequences category, the dataset consists of 36 scenes with a resolution of 1920×1080. The duration ranges from 10 to 487 seconds. The dance sequences involve complex and highly deformable clothing, sports sequences involve interactions between players and props (such as battledore, ball, and taekwondo plank), and kungfu sequences correspond to different choreography types with corresponding costumes and props (such as Broadsword, Nanquan, Nunchuck, Taichi, \etc). In the 4K Studio category, the dataset includes 8 scenes with a resolution of 3840×2160. The duration of each sequence ranges from 10 to 231 seconds. This category covers both solo fashion shows and pair/group performances, as well as basic choreography dances and advanced dances originally choreographed by dancers.

\subsection{Benchmark Pipeline}
The wide range of shapes and textures available in \confDataName~offers a valuable resource for training and evaluating human reconstruction and rendering of both static and dynamic scenes. By leveraging the extensive exploration space offered by \confDataName, we propose an off-the-shelf framework that simplifies the implementation of state-of-the-art NeRF-based methods. This includes neural scene decomposition, 3D human reconstruction, and novel view synthesis of dynamic scenes. Please refer to \cref{fig:pipeline} for more implementation details.



\begin{figure*}[h]
  \centering
   \includegraphics[width=0.98\linewidth]{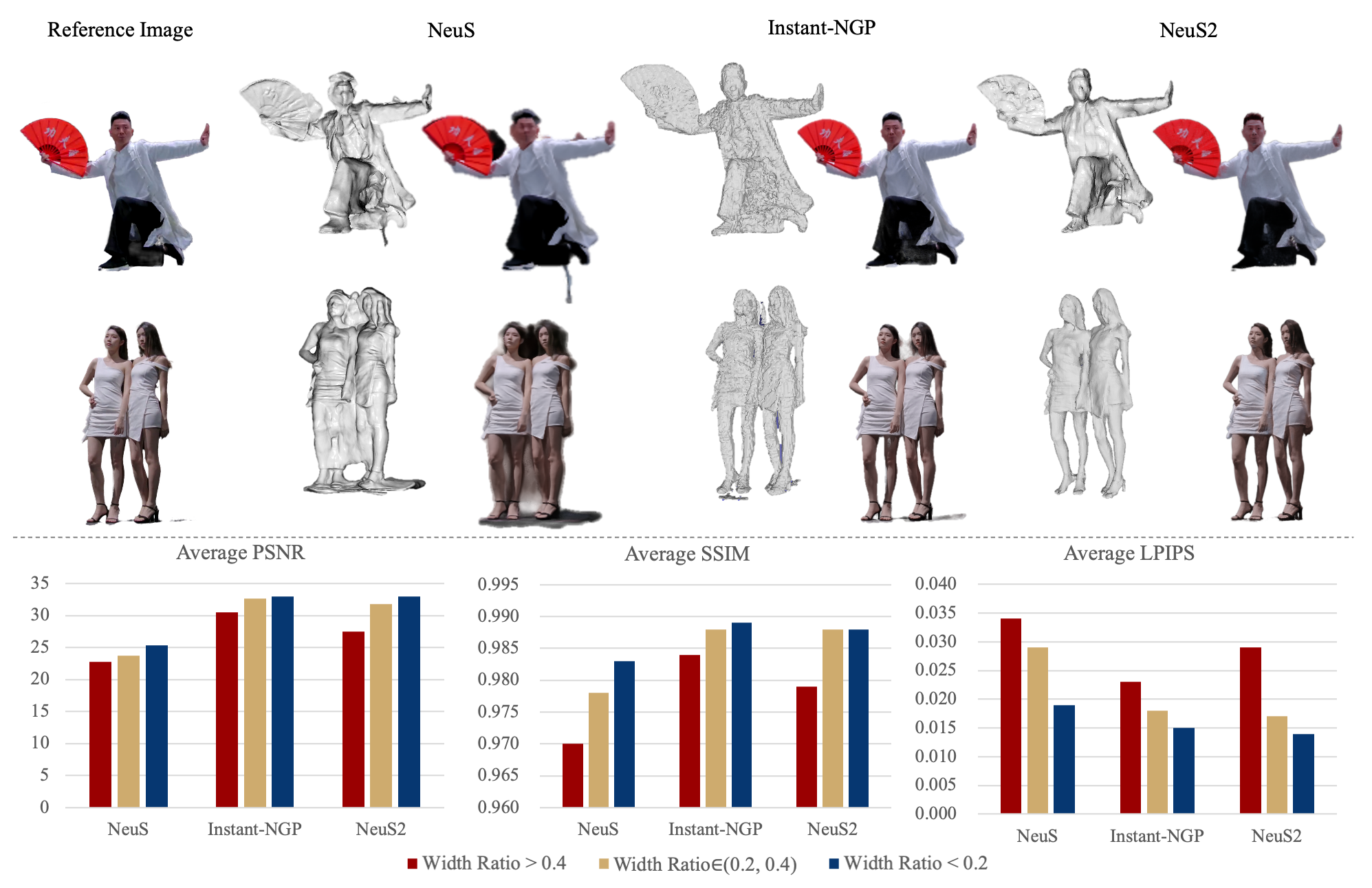}
   \vspace{-2mm}
   \caption{Comparisons on \confDataName~dataset for static scene geometry reconstruction and novel view synthesis.}
   \label{fig:static_rec}
   \vspace{-1mm}
\end{figure*}

\begin{table*}[h]
\scriptsize 
\setlength{\tabcolsep}{5pt} 
\setlength{\extrarowheight}{0.5pt} 
\centering
\caption{Quantitative performance of novel view synthesis on the proposed \confDataName~dataset.}
\vspace{-2mm}
\label{tab:static}
\begin{tabular}{@{}p{14mm}p{38mm}|lll|lll|lll}
\toprule
\multirow{2}{*}{\textbf{Type}}  & \multirow{2}{*}{\textbf{Scenes}} & \multicolumn{3}{c|}{\textbf{NeuS \cite{neus}}}     & \multicolumn{3}{c|}{\textbf{Instant-NGP \cite{instant_ngp}}} & \multicolumn{3}{c}{\textbf{NeuS2 \cite{neus2}}}  \\
                   \textbf{}    &  \textbf{}    & PSNR $\uparrow$ & SSIM $\uparrow$ & LPIPS $\downarrow$  & PSNR $\uparrow$  & SSIM $\uparrow$ & LPIPS $\downarrow$  & PSNR $\uparrow$ & SSIM $\uparrow$ & LPIPS $\downarrow$ \\
\hline
\multirow{4}{14mm}{Width Ratio \textgreater 0.4} & 1080\_Dance\_Dunhuang\_Pair\_f14f15   & 21.38  & 0.959  & 0.042    & 26.37 & 0.974 & 0.035   & 25.34 & 0.967  & 0.044 \\
                                                 & 1080\_Sport\_Taekwondo1\_Pair\_m11c21 & 23.72  & 0.970  & 0.037    & 32.50 & 0.989 & 0.019   & 27.12 & 0.981  & 0.029 \\
                                                 & 1080\_Kungfu\_Sword\_Single\_m13      & 23.24  & 0.978  & 0.023    & 31.10 & 0.986 & 0.019   & 29.39 & 0.985  & 0.019 \\
                                                 & 1080\_Kungfu\_Spear\_Single\_m13      & 22.53  & 0.973  & 0.033    & 31.85 & 0.989 & 0.018   & 28.16 & 0.985  & 0.021  \\
\hline
\multirow{6}{14mm}{Width Ratio $\in(0.2, 0.4)$}  & 1080\_Kungfu\_Fan\_Single\_m12        & 25.72  & 0.981  & 0.024     & 33.16 & 0.988 & 0.021   & 30.56 & 0.985 & 0.021 \\
                                                 & 1080\_Kungfu\_Basic\_Pair\_c24c25     & 25.48  & 0.979  & 0.024     & 31.96 & 0.989 & 0.017   & 30.99 & 0.986 & 0.019 \\
                                                 & 4K\_Studios\_Dance\_Single\_f20       & 22.67  & 0.981  & 0.028     & 30.50 & 0.986 & 0.026   & 31.67 & 0.989 & 0.018 \\
                                                 & 1080\_Dance\_Dunhuang\_Single\_f12    & 25.19  & 0.978  & 0.024     & 31.46 & 0.983 & 0.019   & 30.61 & 0.985 & 0.020 \\
                                                 & 4K\_Studios\_Show\_Single\_f16        & 20.38  & 0.975  & 0.042     & 34.49 & 0.990 & 0.012   & 34.33 & 0.993 & 0.012 \\
                                                 & 4K\_Studios\_Show\_Pair\_f18f19       & 22.80  & 0.976  & 0.031     & 34.24 & 0.993 & 0.016   & 32.23 & 0.992 & 0.014 \\
\hline
\multirow{4}{14mm}{Width Ratio \textless 0.2}    & 1080\_Sport\_Football\_Single\_m11    & 24.91  & 0.983  & 0.017     & 29.83 & 0.982 & 0.018   & 30.50 & 0.986 & 0.016 \\
                                                 & 1080\_Dance\_Banquet\_Single\_c23     & 26.73  & 0.984  & 0.016     & 36.20 & 0.993 & 0.009   & 35.56 & 0.993 & 0.009 \\
                                                 & 1080\_Kungfu\_Tongbeiquan\_Single\_m13 & 23.59  & 0.980  & 0.024    & 32.19 & 0.991 & 0.016   & 31.61 & 0.988 & 0.017 \\
                                                 & 1080\_Dance\_Tibetan\_Single\_c22     & 26.26  & 0.984   & 0.020    & 33.69 & 0.991 & 0.016   & 34.28 & 0.990 & 0.014 \\
\midrule
\multicolumn{2}{c|}{\textbf{Average}}                    & 23.93  & 0.977 & 0.027       & \textbf{32.02} & \textbf{0.987} & \textbf{0.019}       & 30.75 & 0.986 & 0.020 \\
\bottomrule
\end{tabular}
\vspace{-5mm}
\end{table*}

\section{Experiments}
\label{sec:experiments}

\begin{figure*}[h]
  \centering
   \includegraphics[width=0.96\linewidth]{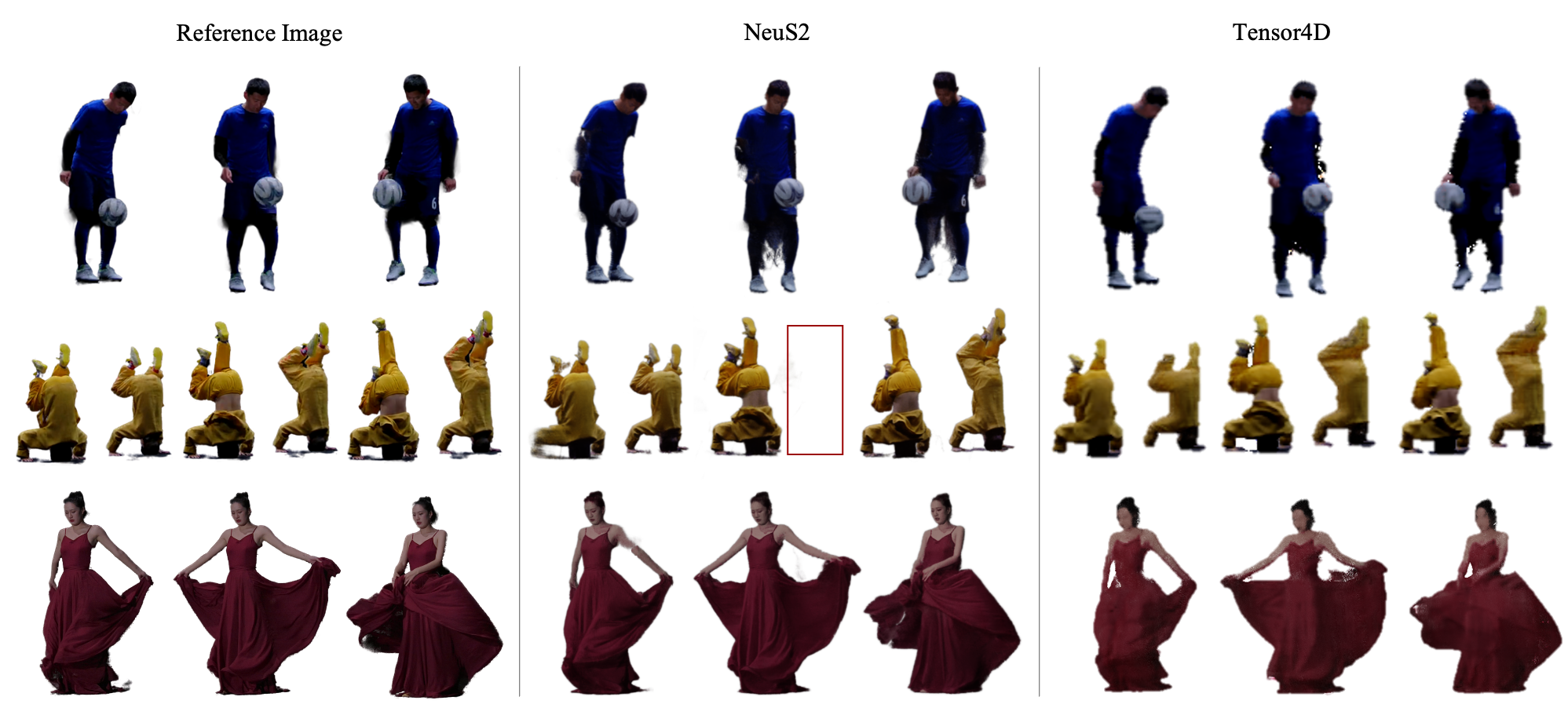}
   
   \vspace{-2mm}
   \caption{The results of space and time novel view rendering under short frame sequences, with dense (NeuS2 \cite{neus2}) and sparse (Tensor4D \cite{tensor4d}) camera views respectively.}
   \label{fig:dyrec}
   \vspace{-5mm}
\end{figure*}

The objective of our benchmark is to achieve robust geometry reconstruction and novel view synthesis for dynamic humans under markerless and fixed multi-view camera settings, while minimizing the need for manual annotation and reducing time costs. To validate the effectiveness and potential of our dataset in neural human modeling, we conduct experiments based on different human width ratios. These experiments cover various action motions, clothing styles, and accessories. Our benchmark focuses on three perspectives, each corresponding to a specific type of application task. These perspectives include multi-view human reconstruction for novel view synthesis, dynamic human modeling, and self-supervised neural scene decomposition. All experiments are conducted on a single GeForce RTX3090 GPU. For more implementation details and additional results, please refer to the supplementary materials.

\subsection{Novel View Synthesis}

{\bf Implementation details.} To evaluate static scene reconstruction, we provide three representative baselines of NeuS \cite{neus}, Instant-NGP \cite{instant_ngp} and NeuS2 \cite{neus2} with \confDataName. All baselines employ publicly available official implementations, and we fine-tune hyperparameters to achieve the best possible results. In the 1080P sequences, we use the same set of 52 training cameras, and 8 validation cameras. For the 4K sequences, we use a set of 48 training cameras and 8 validation cameras. The geometry reconstruction results are generated by training the released code on the training dataset with foreground object mask supervision. We then use the remaining test cameras to obtain novel view synthesis results and calculate metrics such as PSNR, LPIPS \cite{psnr}, and SSIM \cite{ssim}. All metrics are computed over the whole image with white background, and the numerical results are averaged over 6 frames for all experiments. Meanwhile, the scene is divided into three sectors based on the human width ratio. Subsequently, multiple numerical analyses are performed for each sector. 

{\noindent\bf Results analysis.} A qualitative comparison of geometry reconstruction and novel view synthesis results for all methods is presented in \cref{fig:static_rec} and supplementary materials. From the observations in \cref{fig:static_rec}, it is evident that NeuS introduces artifacts when encountering concave geometry (e.g., head, knees, and joints) and suffers from inaccurate density field modeling when the foreground object is dark or in shadow. The extracted meshes of Instant-NGP are noisy due to the lack of surface constraints in the geometry representation. NeuS2 exhibits limited performance with excessively smooth surfaces in the 3D geometry reconstruction. In terms of the novel view synthesis results, although NeuS2 performs better on average, it still lacks visual details and tends to produce blurry results compared to Instant-NGP, as evidenced by distinct cloth wrinkles and folding fan slats.

As shown in \cref{tab:static}, Instant-NGP achieves the highest average performance in terms of PSNR, SSIM, and LPIPS. The best reconstruction achieves a high PSNR of 36.20, indicating that \confDataName~contains the contents that the model can learn and fit well. However, the performance varies, and the lowest PSNR of 20.38 shows that it also includes contents that are outside of the model's learning scope and are challenging. There is a significant gap in these metrics between NeuS and the other two hash encoding-based methods, namely NeuS2 and Instant-NGP, as multi-resolution hash encoding excels at modeling high-frequency details. Additionally, the sector of $Width Ratio \textgreater 0.4$ suffers from lower quality due to complex interactions between humans or humans and objects.

\subsection{Dynamic Human Modeling}

\begin{table*}
\scriptsize 
\setlength{\tabcolsep}{6.1pt} 
\setlength{\extrarowheight}{0.5pt} 
\centering
\caption{Quantitative comparisons between dense and sparse camera views using long and short frame sequences.}
\vspace{-2mm}
\label{tab:dyrec}
\begin{tabular}{p{28mm}|p{31mm}<{\centering}|p{30mm}<{\centering}|p{30mm}<{\centering}|p{31mm}<{\centering}}
\toprule
\multirow{3}{*}{\textbf{Width ratio of human box}} & {\textbf{Instant-NGP \cite{instant_ngp}}} & {\textbf{NeuS2 \cite{neus2}}} & {\textbf{NeuS2 \cite{neus2}}}        & {\textbf{Tensor4D \cite{tensor4d}}}  \\ 
                 & {\textbf{Long sequence(Dense-view)}}  & {\textbf{Long sequence(Dense-view)}}  & {\textbf{Short sequence(Dense-view)}} & {\textbf{Short sequence(Sparse-view)}}    \\ 
 & PSNR $\uparrow$ \// SSIM $\uparrow$ \// LPIPS $\downarrow$   & PSNR $\uparrow$ \// SSIM $\uparrow$ \// LPIPS $\downarrow$     & PSNR $\uparrow$ \// SSIM $\uparrow$ \// LPIPS $\downarrow$       & PSNR $\uparrow$ \// SSIM $\uparrow$ \// LPIPS $\downarrow$  \\
\midrule
Width Ratio \textgreater 0.4  & 29.53 \// 0.984 \// 0.025       & 25.87 \// 0.974 \// 0.032        & 28.62 \// 0.980 \// 0.025    & 27.59 \// 0.970 \// 0.040 \\
Width Ratio $\in(0.2, 0.4)$   & 30.55 \// 0.983 \// 0.022       & 30.05 \// 0.985 \// 0.021        & 32.04 \// 0.989 \// 0.017    & 31.56 \// 0.982 \// 0.028\\
Width Ratio \textless 0.2     & 29.39 \// 0.984 \// 0.022       & 30.52 \// 0.987 \// 0.018        & 32.93 \// 0.989 \// 0.015    & 28.89  \// 0.986 \// 0.022   \\
\midrule
\textbf{Average}              & 29.82 \// 0.984 \// 0.023       & 28.81 \// 0.982 \// 0.024       & \textbf{31.20 \// 0.986 \// 0.019}  & 29.38 \// 0.979 \// 0.030 \\
\bottomrule
\end{tabular}

\vspace{-4mm}
\end{table*}

{\bf Implementation details.} We apply the task of dynamic human reconstruction with two different input settings: 1) Dynamic reconstruction under densely captured videos: In this setting, each sequence contains 10 to 500 frames with 48 or 52 camera views for training and 8 camera views for testing. Given the multi-view videos of a moving object and camera parameters for each view, we use Instant-NGP and NeuS2 to recover per-frame reconstruction and dynamic reconstruction, respectively. 2) Dynamic reconstruction using sparse and fixed cameras: We use 16 cameras focusing on the front face, with the number of training frames ranging from 10 to 20. For this setting, we employ NeuS2 and Tensor4D to represent topologically varying objects. For quantitative evaluation, we calculate and average the PSNR, LPIPS, and SSIM scores over all frames.

{\noindent\bf Results analysis.} We present example results for novel view synthesis in \cref{fig:dyrec}. As shown in \cref{fig:dyrec}, when dealing with long sequences of 10 frames that involve challenging movements or multiple foreground objects, NeuS2 struggles to accurately reconstruct dynamic objects. Similarly, for sparse-view videos, Tensor4D struggles to render high-quality images for dynamic scenes and faithfully recover appearance details such as thin finger motions, human-object interaction, facial expressions, and cloth wrinkles. 

The quantitative results are summarized in \cref{tab:dyrec}, which include both familiar contents that the model can handle well and more challenging new contents, demonstrating the diversity of our dataset. Additionally, we observe that the diverse motions, appearances, long sequences, and loose garments in \confDataName~pose further challenges for 4D neural human rendering.


\subsection{Neural Scene Decomposition}
\label{subsec:nsd}
Neural scene decomposition aims to effectively separate the foreground and background without any annotations, with downstream applications in multi-view object segmentation \cite{view_cons} and beyond, such as 3D/4D human reconstruction and rendering. Under conditions of multi-camera inputs, the structural, textural and geometrical consistency among each view can be leveraged to achieve fine-grained object segmentation. Surface-SOS \cite{sos} addresses this information by combining the neural surface representation from multi-view observations via volume rendering of SDF. It captures the compositional nature of scenes and provides additional inherent information, improving 3D human reconstruction.

\begin{figure}
  \centering
   \includegraphics[width=0.96\linewidth]{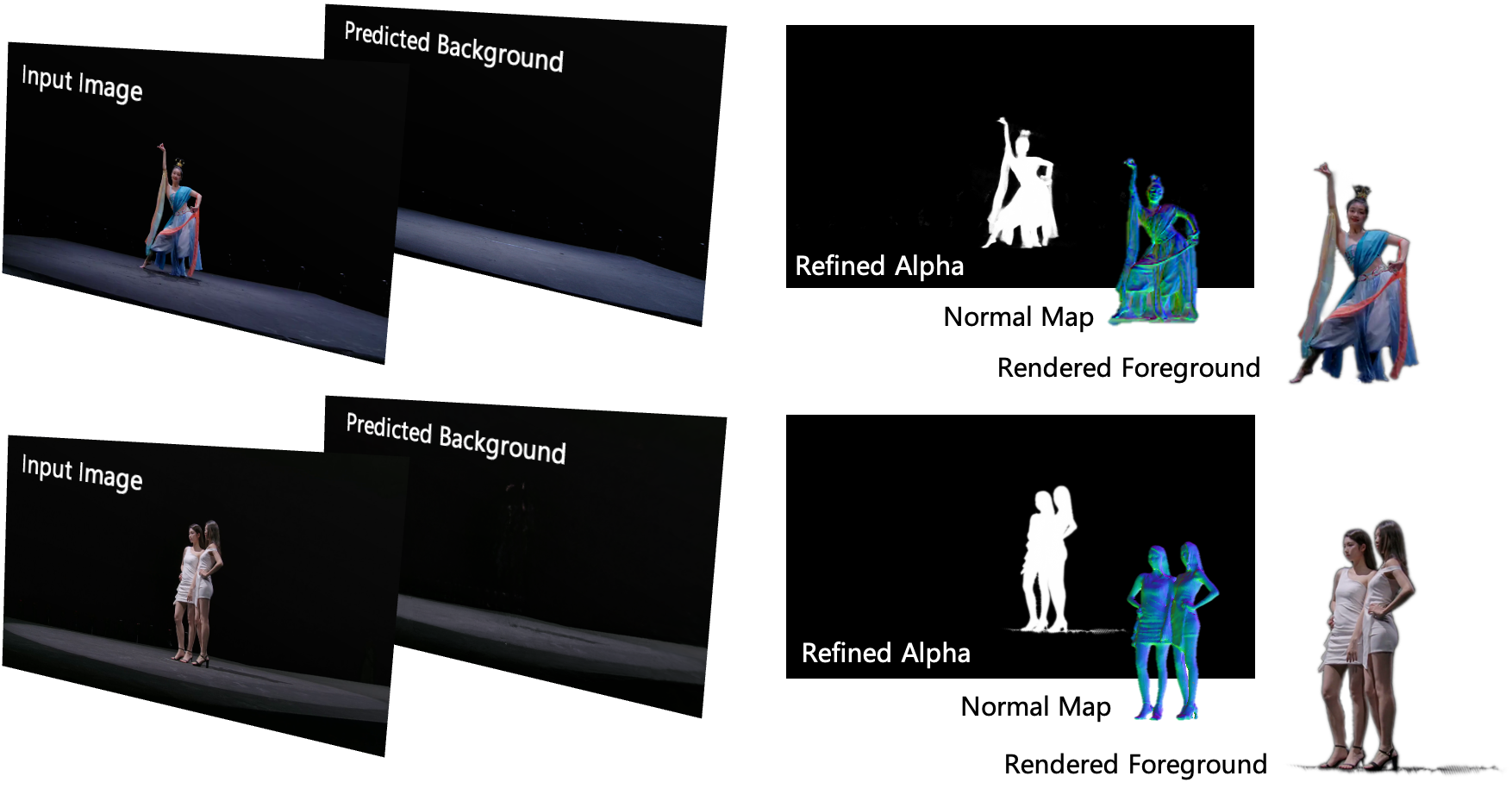}
   
   \vspace{-2mm}
   \caption{Given multi-view images as input, the goal is to decompose 3D scenes into geometrically consistent foreground objects, texture-completed backgrounds, and generate convincing segmentation and normal maps for them.}
   \label{fig:seg}
   \vspace{-5mm}
\end{figure}

We train Surface-SOS \cite{sos} for each individual sequence and present some example results in \cref{fig:seg}. It shows that Surface-SOS can generates detailed foreground alpha, resulting in high-quality geometry surfaces. This is evidenced by the distinct cloth wrinkles and natural shading. The quantitative results in \cref{tab:nsd} also highlights the improvement of the Surface-SOS refinement.
The detailed procedures and more comparison result against relevant methods of \confDataName~are further illustrated in the supplementary material. It shows that when dealing with scenes involving human-object interactions, multi-person interactions, as well as challenging complex scene effects, it faces even greater challenges. 

\begin{table}
\scriptsize 
\setlength{\tabcolsep}{4pt} 
\setlength{\extrarowheight}{0.35pt} 
\centering
\caption{Quantitative evaluation of the foreground segmentation on \confDataName~dataset.}
\vspace{-2mm}
\label{tab:nsd}
\begin{tabular}{lcc}
\toprule
\multirow{2}{*}{\textbf{Width ratio of human box}} & {\textbf{BGMv2 \cite{bgmv2}}}    & {\textbf{Surface-SOS \cite{sos}}} \\ 
                        & MSE$\downarrow$ \// mIoU$\uparrow$ \// Acc.$\uparrow$       & MSE$\downarrow$ \// mIoU$\uparrow$ \// Acc.$\uparrow$ \\
\midrule
Width Ratio \textgreater 0.4                      & 0.345 \// 0.853 \// 0.925         & 0.226 \// 0.871 \// 0.943   \\
Width Ratio $\in(0.2, 0.4)$                       & 0.482 \// 0.891 \// 0.932         & 0.219 \// 0.936 \// 0.970   \\
Width Ratio \textless 0.2                         & 0.439 \// 0.900 \// 0.958         & 0.228 \// 0.917 \// 0.973  \\
\midrule
\textbf{Average}                                  & 0.422 \// 0.881 \// 0.938         & \textbf{0.224 \// 0.908 \// 0.962} \\
\bottomrule
\end{tabular}

\vspace{-3mm}
\end{table}

\section{Conclusion}
In this work, we introduced \confDataName, a dynamic human dataset designed for high-fidelity human reconstruction and rendering from dense multi-view videos. It features high-fidelity human performance, including high-detailed appearance, complex human motion, as well as challenging human-object interactions, multi-person interactions and complex scene effects (\eg, lighting, shadows, and smoking). We further presented benchmark tasks, with detailed experiments on several advanced methods. \confDataName~further fill in the gap between existing datasets and real-scene applications. 


{\bf Challenges and future works.} While we have validated the complexity and fidelity of our dataset on numerous human-centric reconstruction and rendering scenarios. It is significant to highlight the more challenging and realistic multiple-person/subject modeling that could reflect the rendering differences with respect to multi-person interactivity, complex scene effects, and multi-view consistent performance.
Additionally, free-viewpoint rendering of a moving subject from a monocular self-rotating video is a complex yet desirable setup. Our supplementary material provides additional experiments for free-viewpoint rendering of moving subjects, the results are affected by local occlusion and view absence, leading to artifacts in view rendering.
With these opportunities and challenges, we believe \confDataName~will benefit the development of new approaches in the community.

{\bf Acknowledgments.} This work is supported by Outstanding Talents Training Fund in Shenzhen, Shenzhen Science and Technology Program (RCJC20200714114435057, SGDX20211123144400001), National Natural Science Foundation of China (U21B2012), and Migu-PKU Meta Vision Technology Innovation Laboratory (R24115SG). Jianbo Jiao is supported by the Royal Society grants IES\textbackslash R3\textbackslash223050 and SIF\textbackslash R1\textbackslash231009.

\newpage

{
    \small
    \bibliographystyle{ieeenat_fullname}
    \bibliography{main}
}

\clearpage
\maketitlesupplementary
This supplementary material presents more details and additional results not included in the main paper due to page limitation. The list of items included are: 

\begin{itemize}
\item More dataset information in Sec.~\ref{sec:info}, including visualizations of data samples, and per-category data distribution;

\item Additional experimental details are provided in Sec.~\ref{sec:expri}. These include implementation details of neural scene decomposition, additional visualizations and quantitative results of novel view synthesis, as well as more evaluation results on dynamic human modeling.
\end{itemize}
  
\appendix
\section{Additional Dataset Information}
\label{sec:info}

\subsection{More Visualizations of Data Samples}

\begin{figure*}[b]
  \centering
   \includegraphics[width=\linewidth]{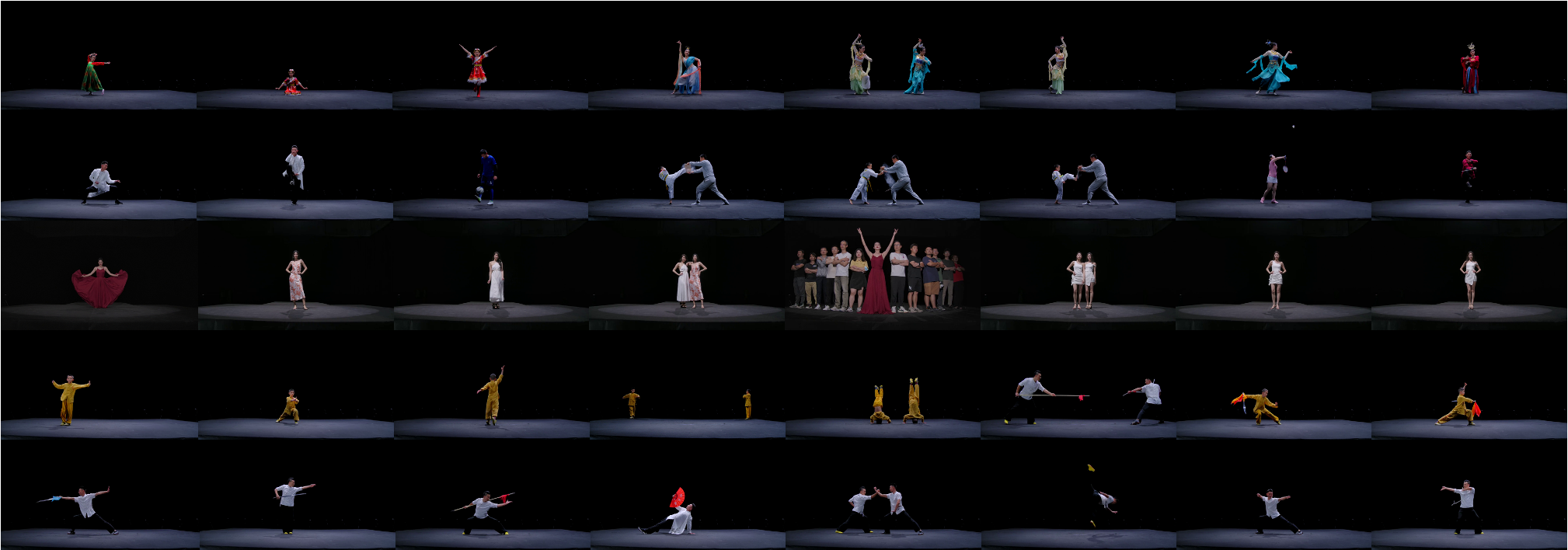}
   \caption{\textbf{Data overview}. \confDataName~is a dynamic, human-centric dataset with diverse subjects, each featuring highly detailed appearances and complex human motions.}
   \label{fig:overview}
\end{figure*}

\begin{figure*}[b]
  \centering
   \includegraphics[width=\linewidth]{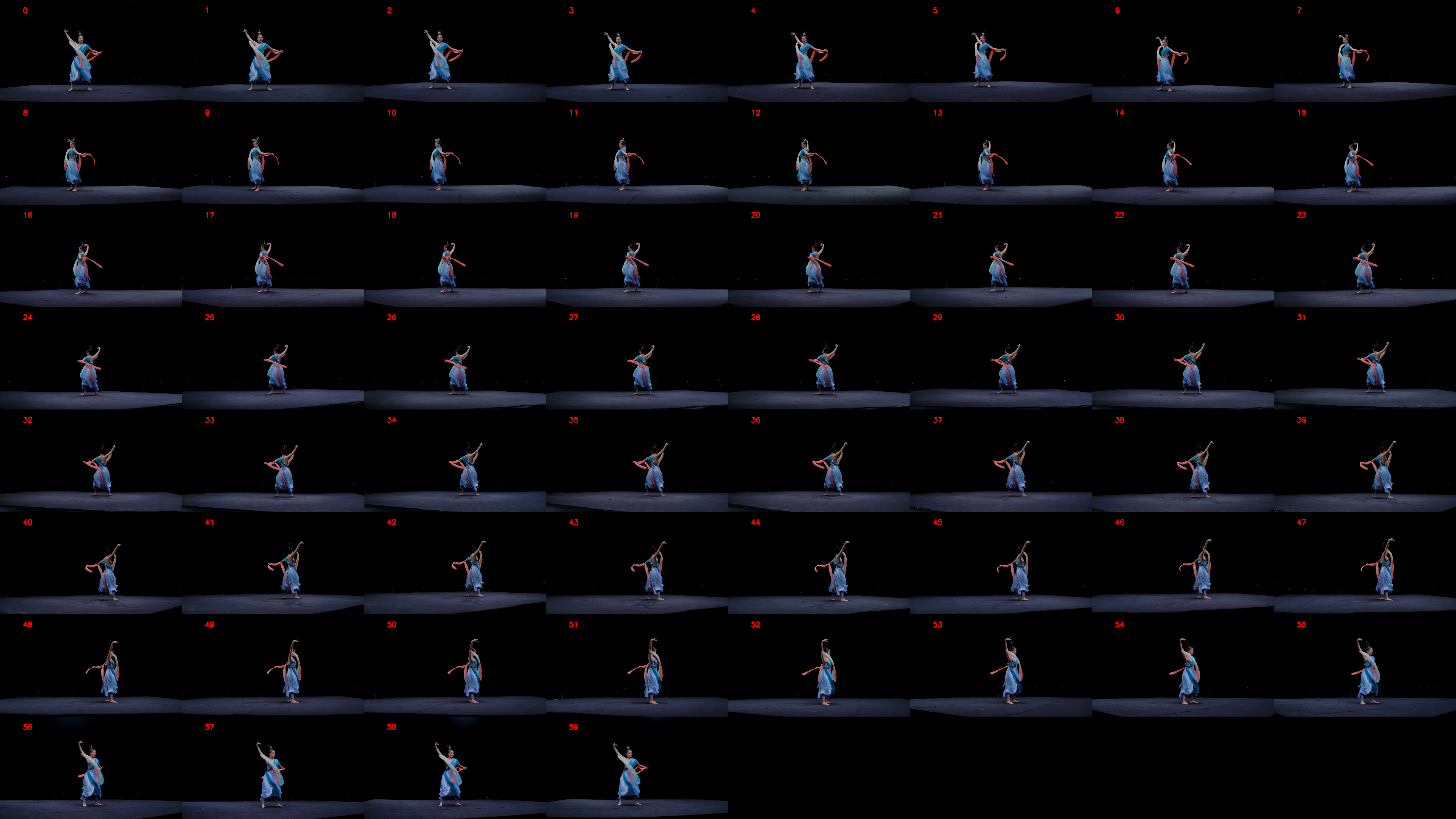}
   \caption{A set of example multi-view images in the 1080P sequences (1080\_Dance\_Dunhuang\_Single\_f12).}
   \label{fig:sample1}
\end{figure*}

\begin{figure*}[b]
  \centering
   \includegraphics[width=\linewidth]{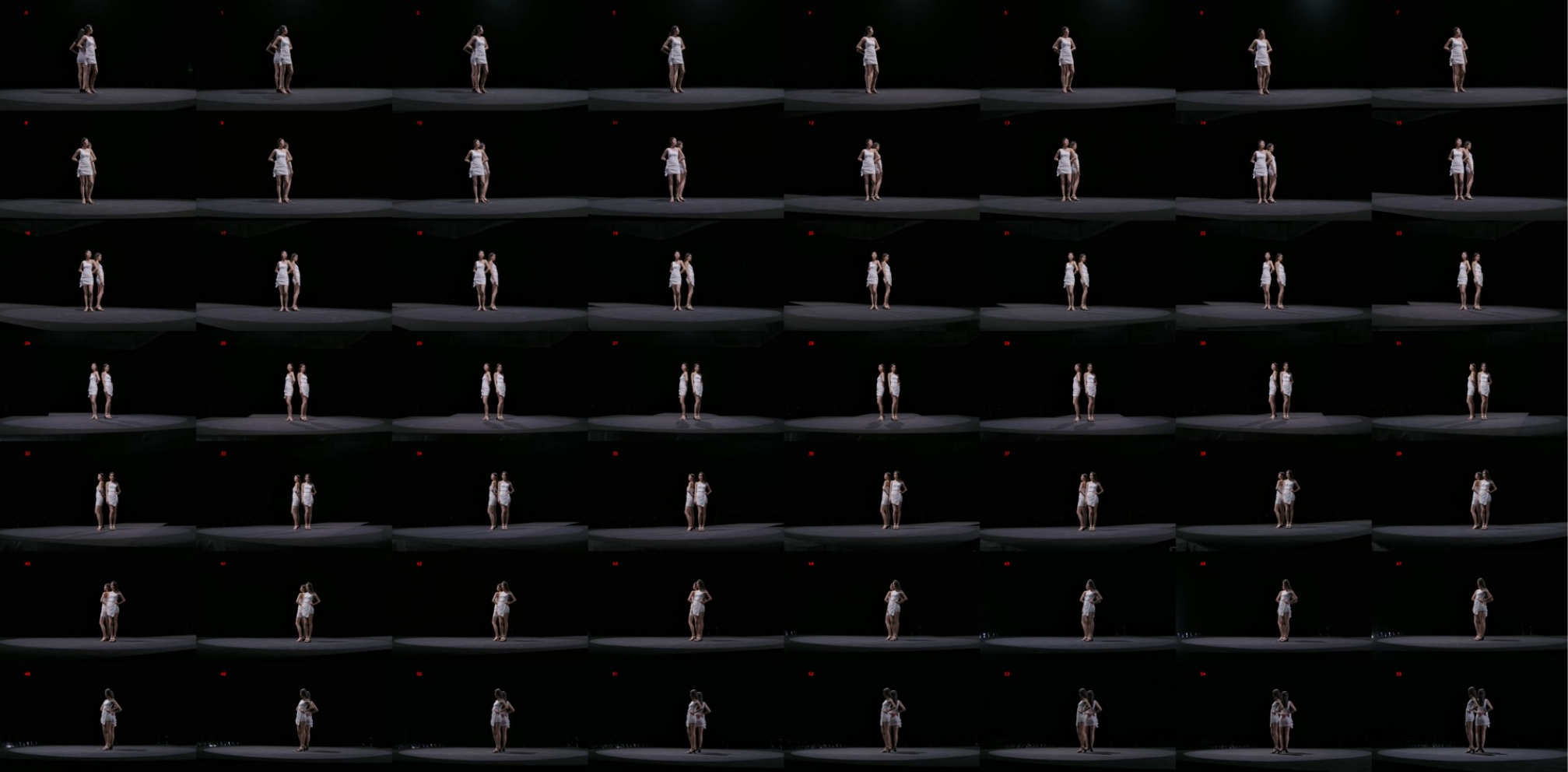}
   \caption{A set of example multi-view images in the 4K Studio sequences (4K\_Studios\_Show\_Pair\_f18f19).}
   \label{fig:sample2}
\end{figure*}

\cref{fig:overview} presents an overview of the sample for each scene in \confDataName. It can be seen that each sample has a distinct texture, motion, and interactions, covering a wide range of fundamental and complex dynamic performances. As shown in \cref{fig:sample1} and \cref{fig:sample2}, a set of multi-view images is illustrated for the 1080P and 4K Studio sequences category, respectively. It clearly shows the differences between each view and provides comprehensive categories in our dataset.

\subsection{Per-category Data Distribution}

\begin{figure*}[b]
  \centering
   \includegraphics[width=\linewidth]{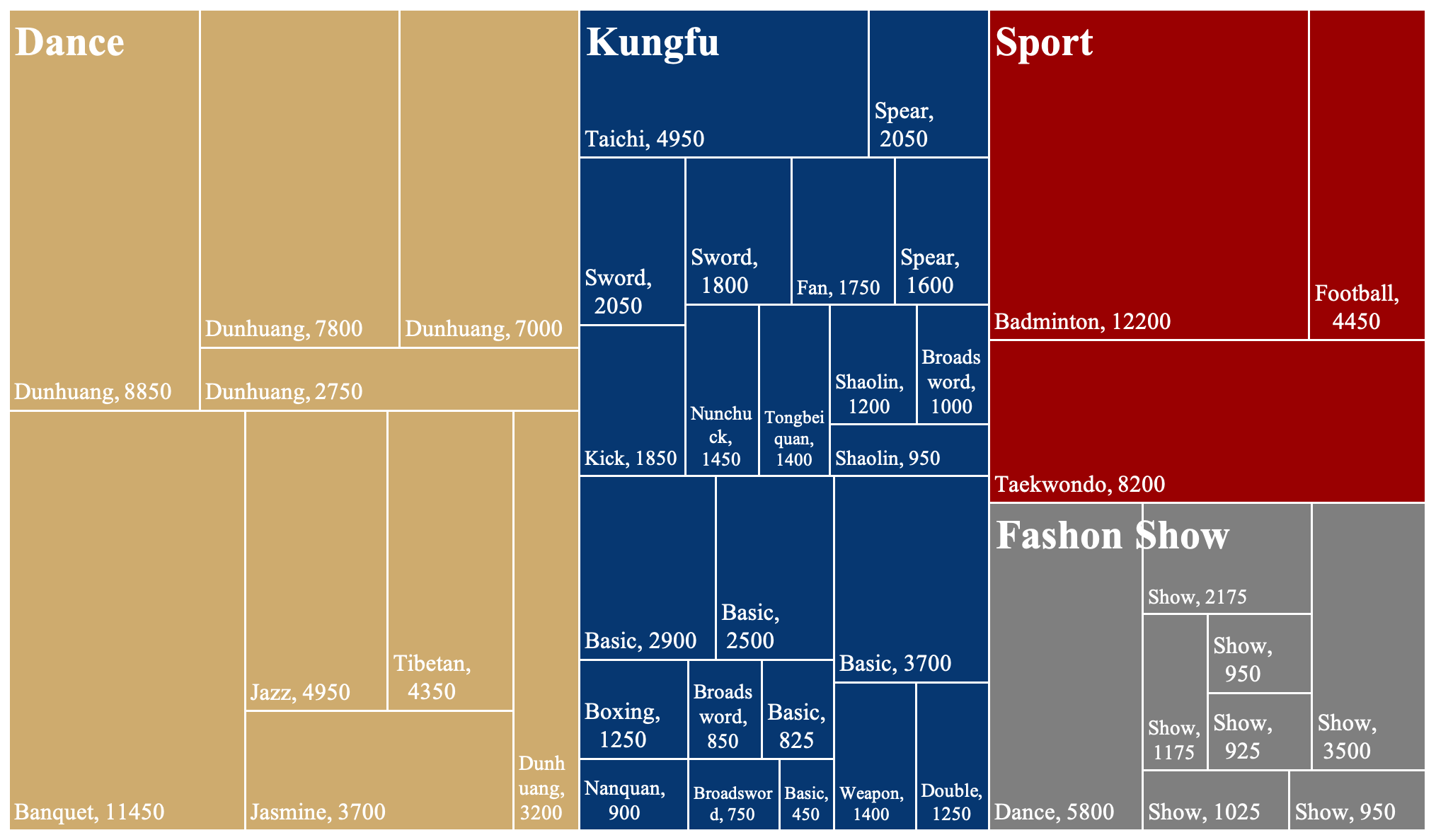}
   \caption{A scene list that provides a detailed breakdown of the number of frames in each category.}
   \label{fig:fme_distri}
\end{figure*}

\confDataName~contains a massive scale of subjects (32), scene (45), sequences (2,668) and frames (8.2M). We provide a full scene distribution with the number of frames for each action category in \cref{fig:fme_distri}. 

\section{Additional Experimental Details}
\label{sec:expri}

\subsection{Neural Scene Decomposition}
\begin{figure*}[h]
  \centering
   \includegraphics[width=0.98\linewidth]{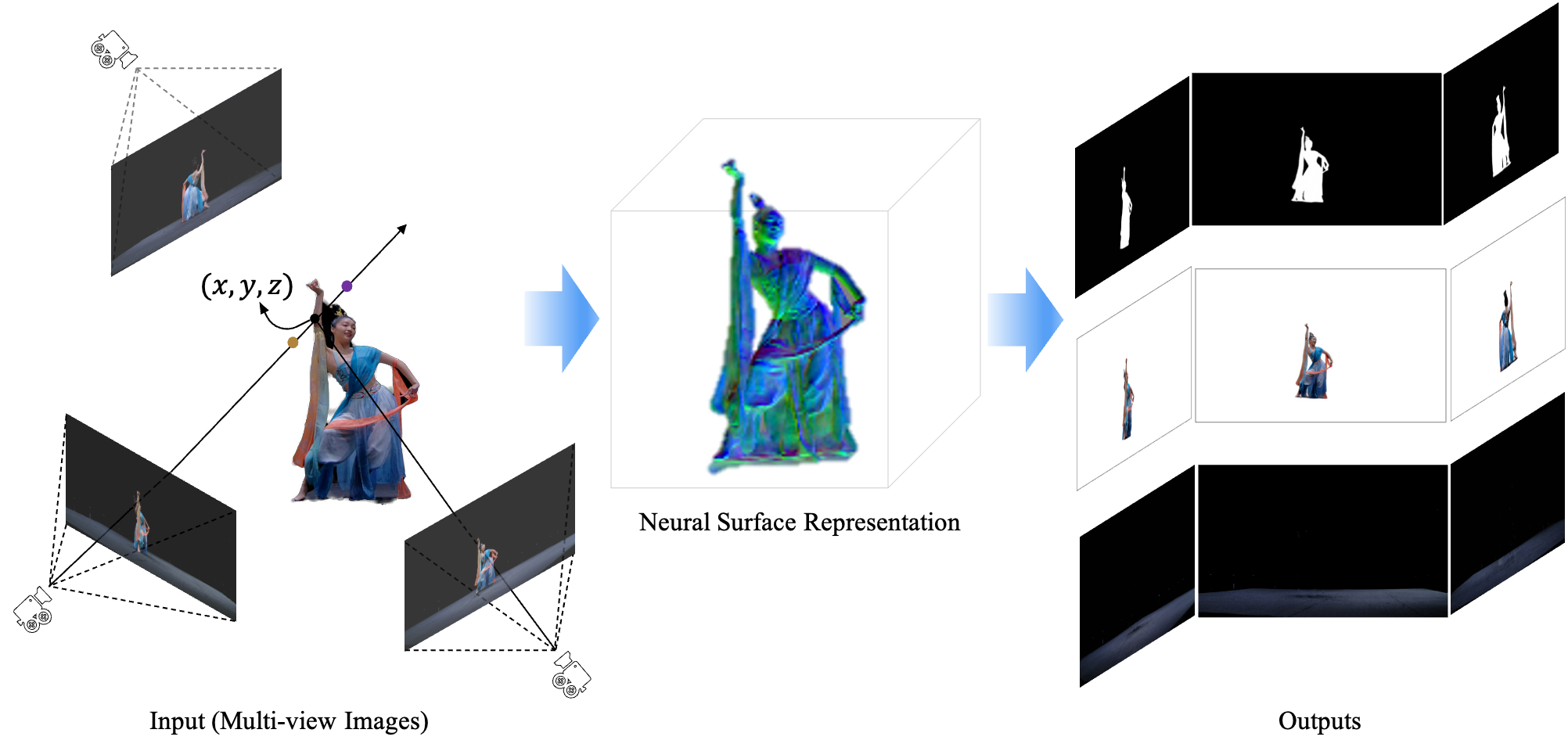}
   \caption{Surface-SOS \cite{sos}, a self-supervised learning framework towards delicate segmentation by combining 3D neural surface representation power from multi-view images of a scene.}
   \label{fig:sos}
\end{figure*}

{\bf Method overview.} For the scene captured by multi-view images, we use COLMAP \cite{sfm} and BGMv2 \cite{bgmv2} to get sparse 3D points and coarse object masks as co-inputs, and predict a dense, geometrical consistent object map, as well as a textural, completed background for each image. \cref{fig:sos} shows an overview of our contemporary work Surface-SOS \cite{sos}, in which multi-view geometric constraints are embedded in the form of dense one-to-one mapping in 3D surface representation. By connecting SDF-based surface representation to geometric consistency, and applying volume rendering to train the network with robustness, it can reconstruct the foreground object geometry and appearance over time.

{\bf Implementation details.} To train Surface-SOS \cite{sos}, we introduce photometric and geometric losses to supervise the training process, with the multi-view images serving as the primary supervision signal. Our objective is to achieve fine-grained object segmentation and analyze the correlation between the neural surface representation and object segmentation. In this study, we evaluate two approaches: NeRF-based segmentation, which does not introduce the normal to regularize the output SDF implicitly, and SDF-based segmentation, which provides the SDF-based surface representation for cross-view geometric constraints. 

\begin{figure*}[h]
  \centering
   \includegraphics[width=\linewidth]{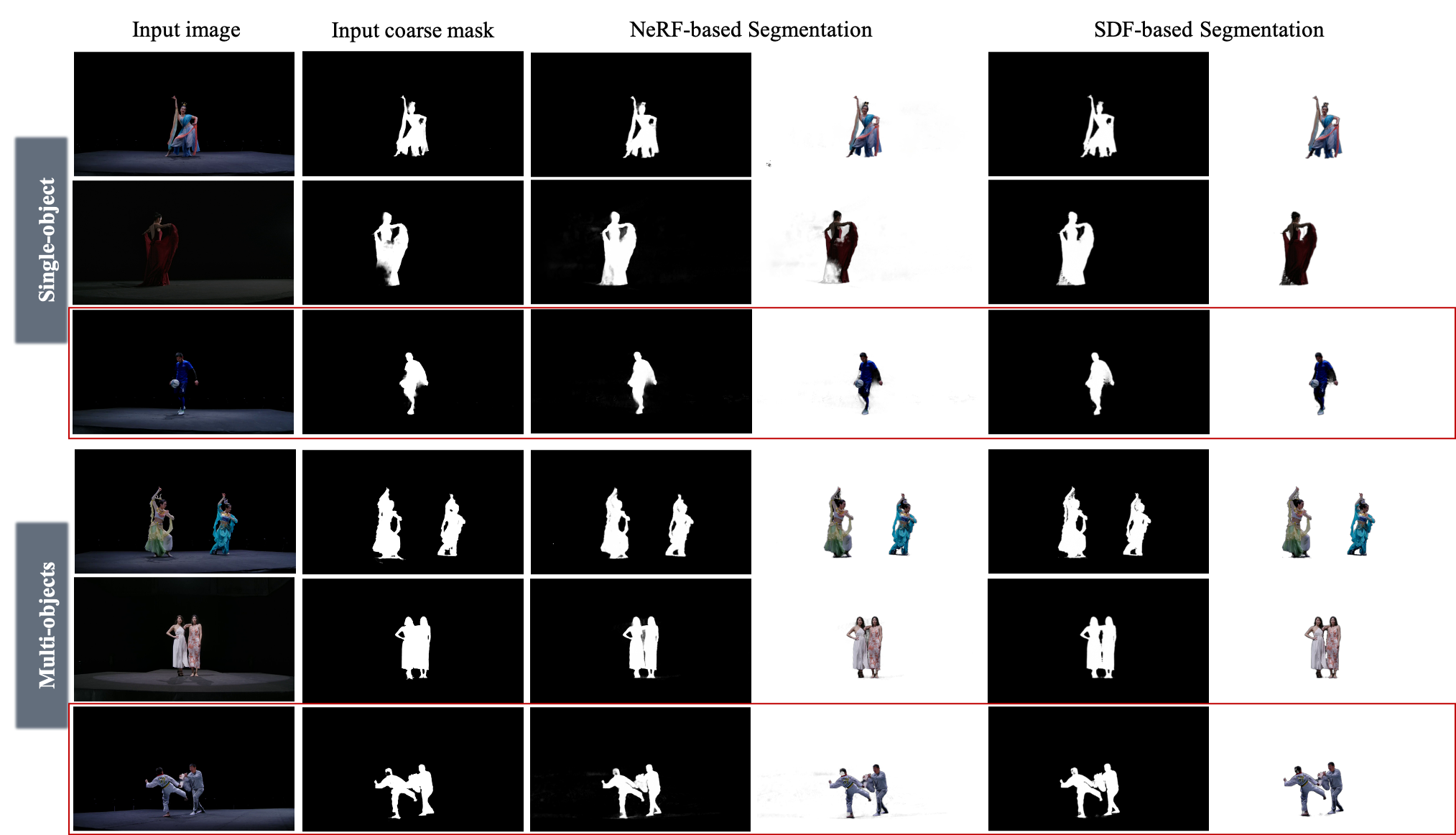}
   \caption{Qualitative comparison on 3D segmentation of scenes with a single/multiple foreground object.}
   \label{fig:decomp}
\end{figure*}

{\bf 3D Segmentation of scenes with a single/multiple foreground object.}  As shown in \cref{fig:decomp}, Surface-SOS successfully refines the segmentation remarkably using two neural representation models. When the normal is not introduced to implicitly regularize the output SDF (i.e., NeRF-based segmentation), it often produces noisy segmentation. However, when providing the SDF-based surface representation, the network is able to learn 3D geometry implicitly and generate an accurate foreground decomposition. These examples demonstrate that accurate prediction of object geometry with SDF-based surface representation is beneficial for object segmentation. 

\subsection{Novel View Synthesis}
\begin{figure*}[h]
  \centering
   \includegraphics[width=\linewidth]{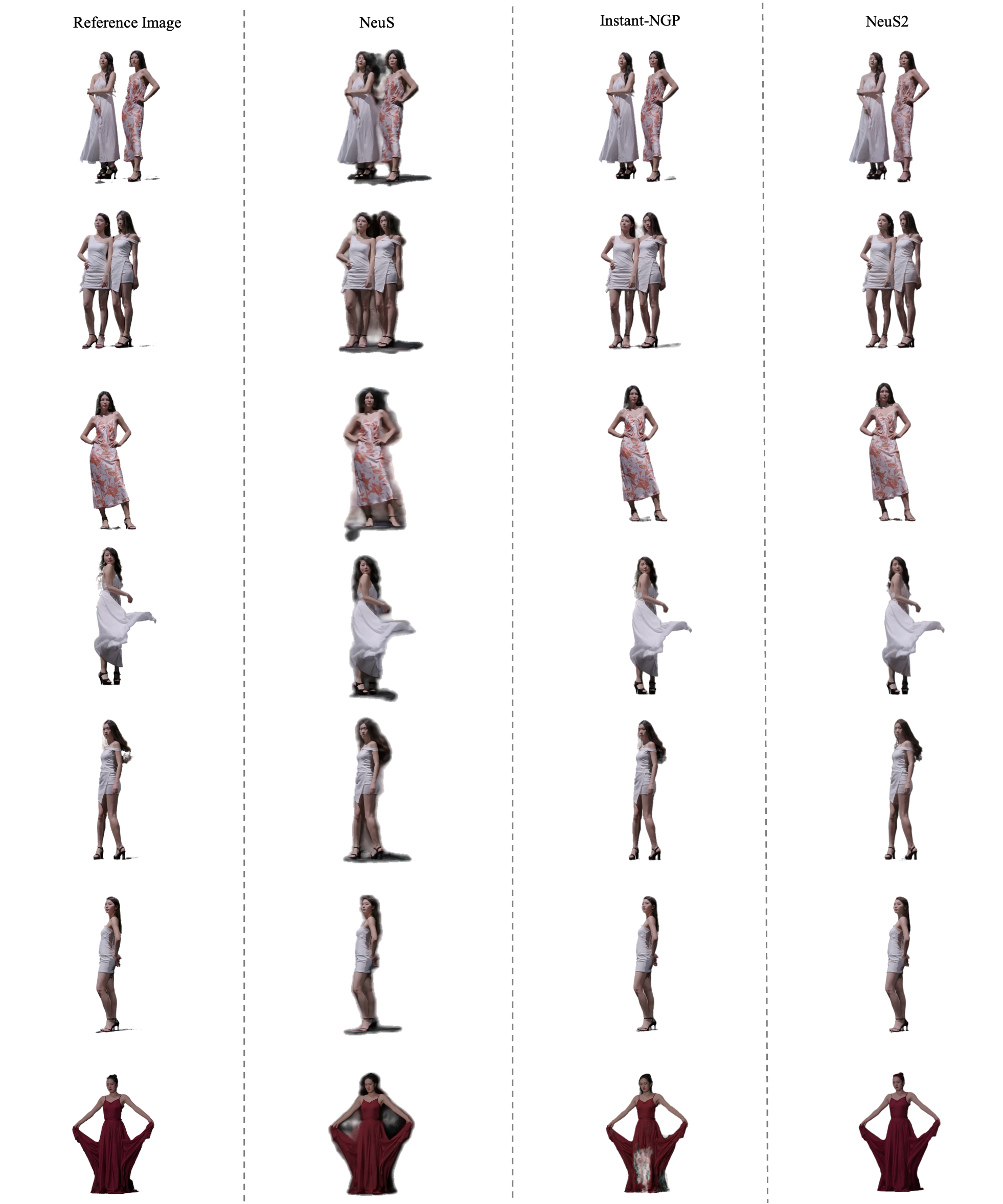}
   \caption{More visualizations results of scene sample (4K Studios).}
   \label{fig:nvs_4k}
\end{figure*}

\begin{figure*}[h]
  \centering
   \includegraphics[width=\linewidth]{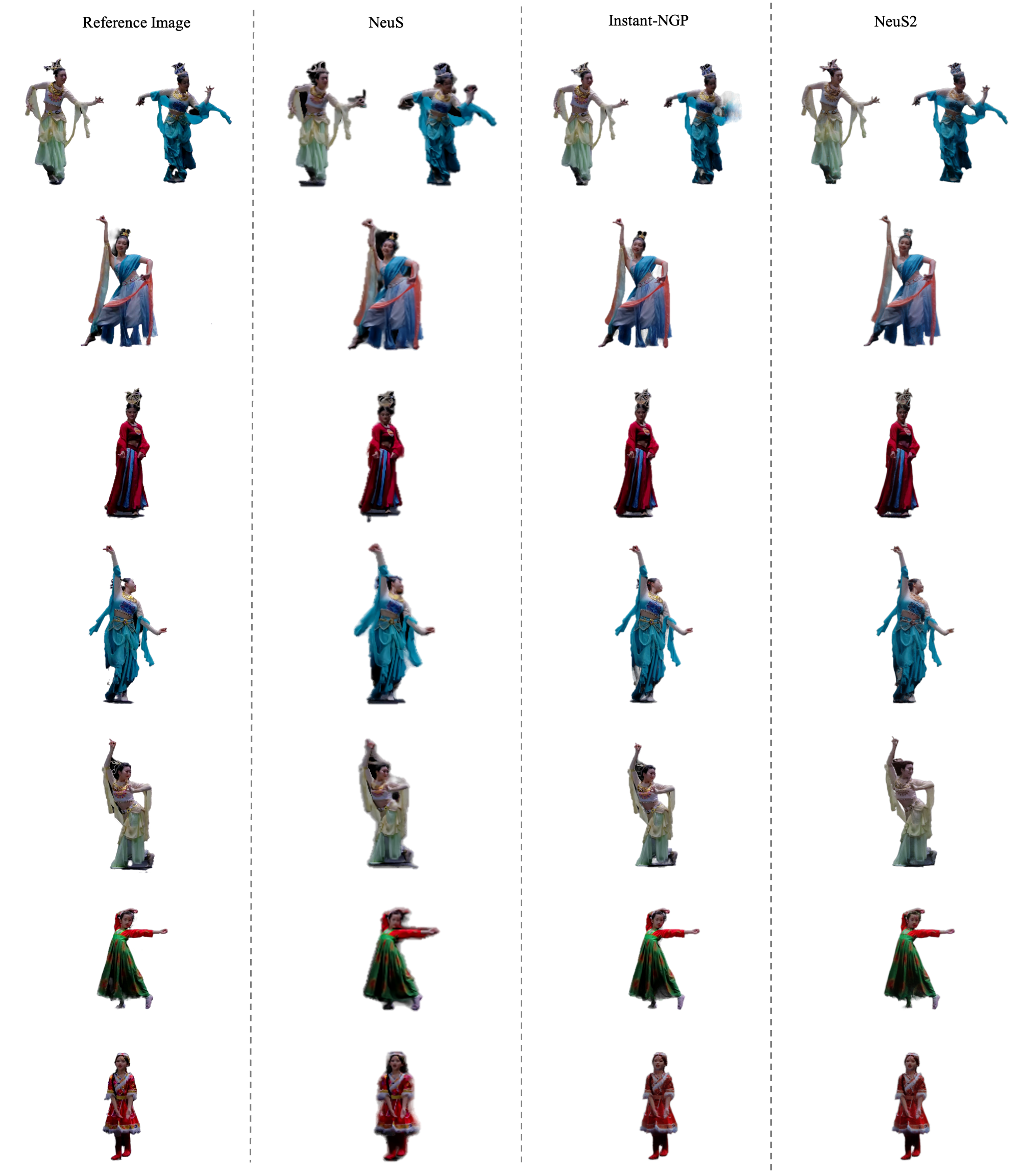}
   \caption{More visualizations results of scene sample (dance).}
   \label{fig:nvs_dance}
\end{figure*}

\begin{figure*}[h]
  \centering
   \includegraphics[width=\linewidth]{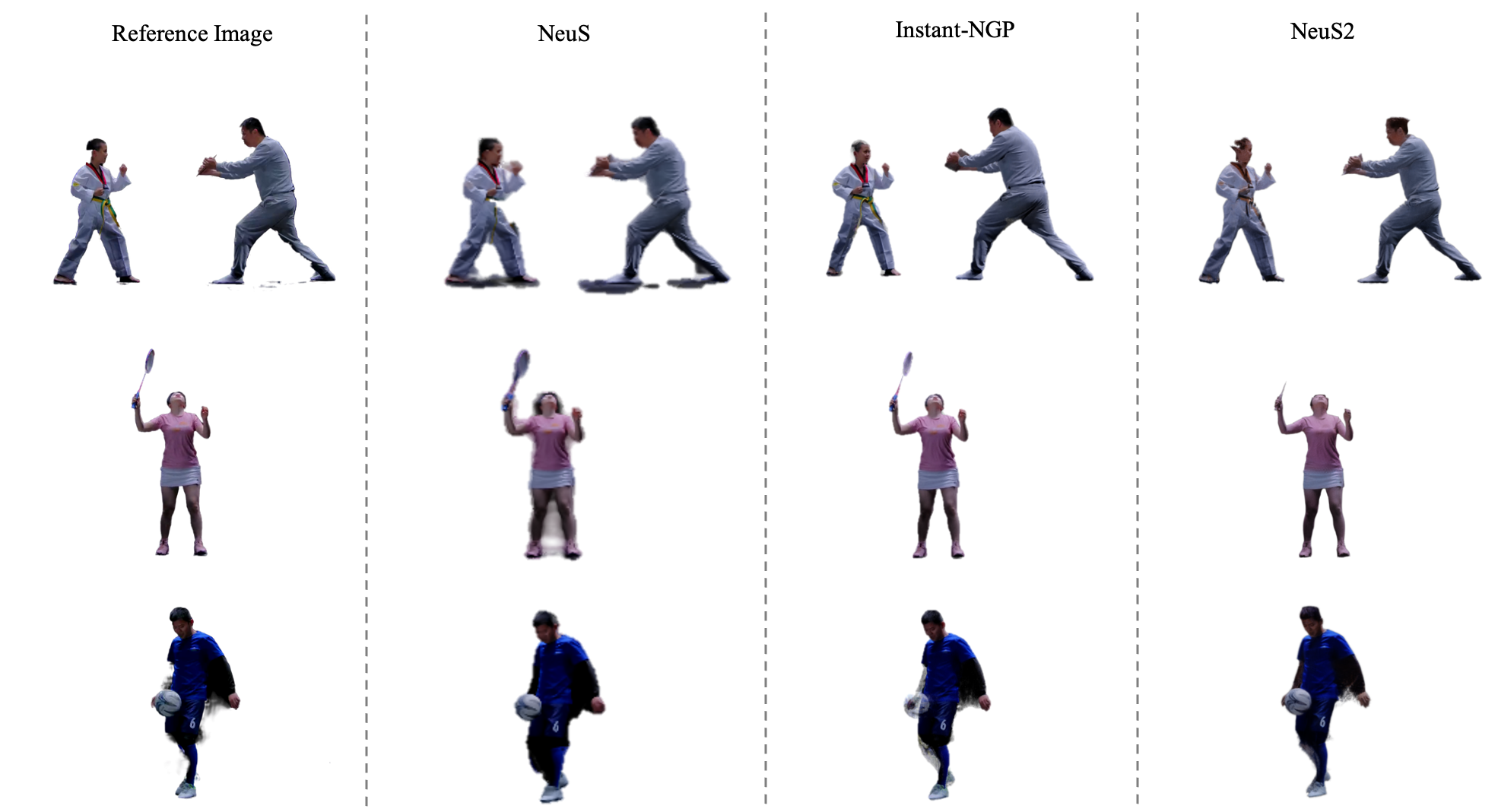}
   \caption{More visualizations results of scene sample (sport).}
   \label{fig:nvs_sport}
\end{figure*}

\begin{figure*}[h]
  \centering
   \includegraphics[width=\linewidth]{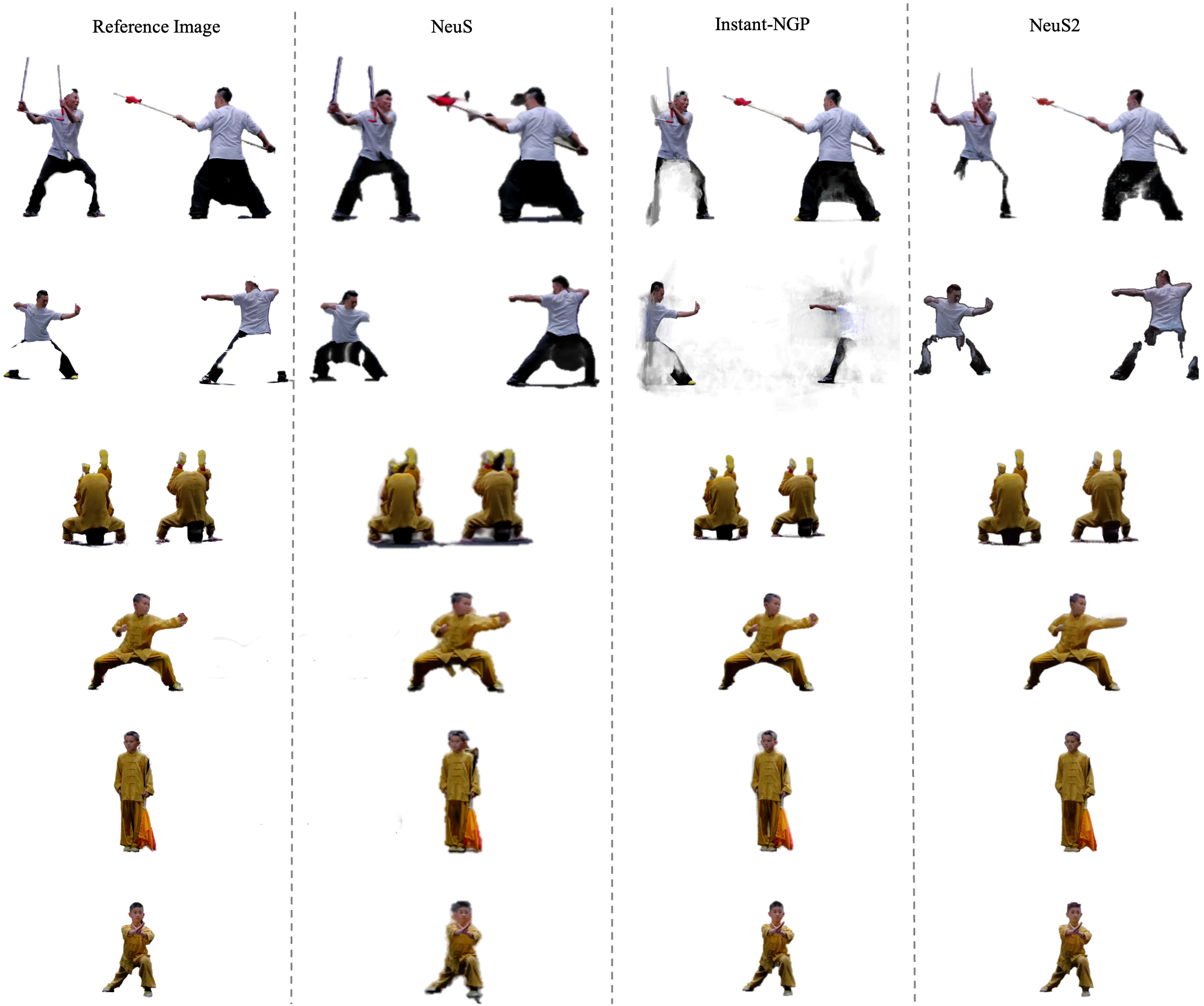}
   \caption{More visualizations results of scene sample (kungfu).}
   \label{fig:nvs_kufung1}
\end{figure*}

\begin{figure*}[h]
  \centering
   \includegraphics[width=\linewidth]{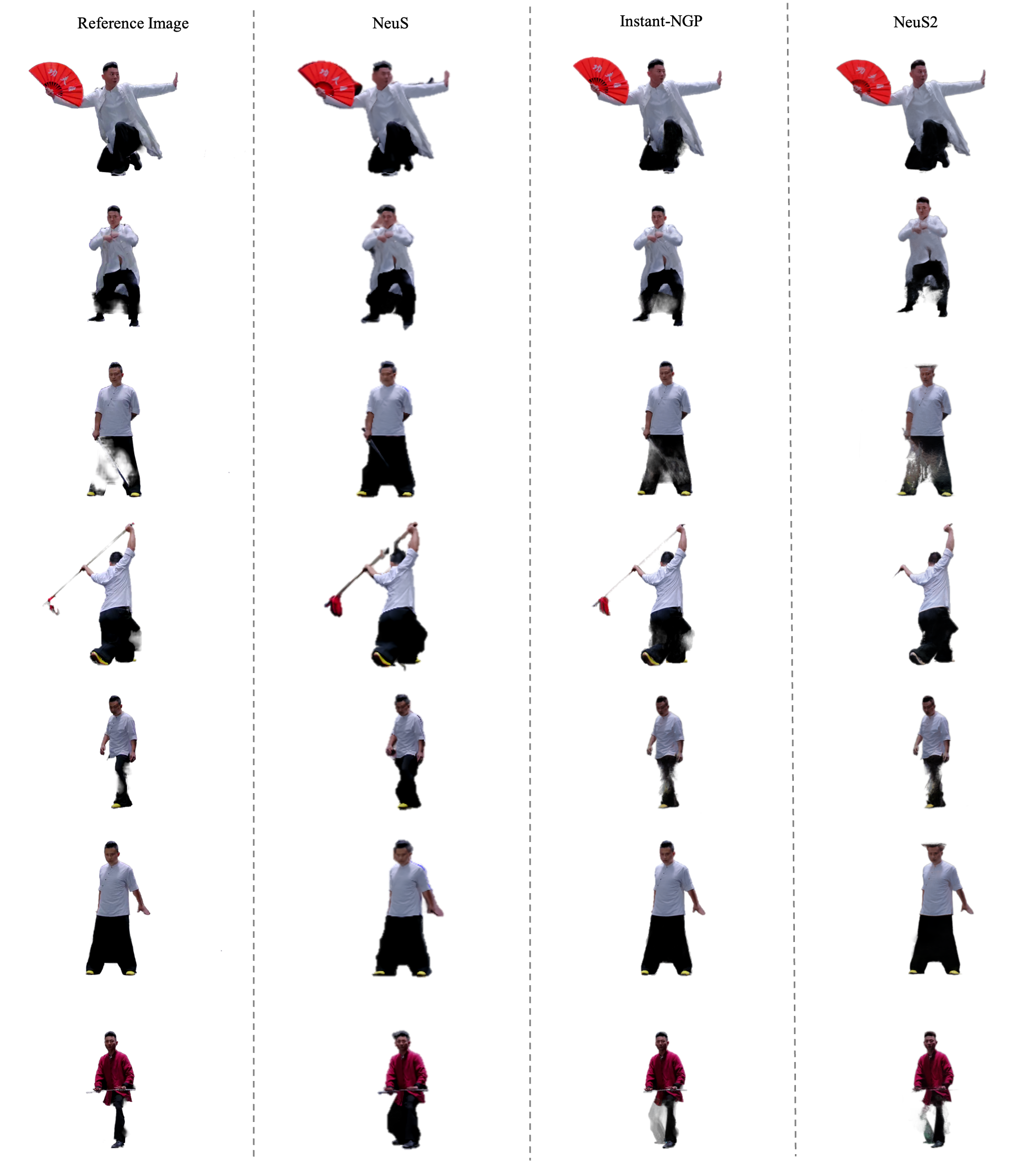}
   \caption{More visualizations results of scene sample (kungfu).}
   \label{fig:nvs_kufung2}
\end{figure*}

We conducted an additional experiment on the remaining scenes in \confDataName~dataset. The complete quantitative comparisons are presented in \cref{tab:nvs}. Additionally, we present additional qualitative comparisons in \cref{fig:nvs_4k}, \cref{fig:nvs_dance}, \cref{fig:nvs_sport}, \cref{fig:nvs_kufung1}, and \cref{fig:nvs_kufung2}. Consistent with the results in the main paper, \confDataName~dataset offers a wide range of shapes and appearances, providing a comprehensive foundation for evaluating various methods for novel view synthesis in terms of human performance.


\begin{table*}
\scriptsize 
\centering
\caption{Results of per-scene novel view synthesis on 4 action categories.}
\vspace{-2mm}
\label{tab:nvs}
\begin{tabular}{@{}p{14mm}p{38mm}|lll|lll|lll}
\toprule
\multirow{2}{*}{\textbf{Action Type}}  & \multirow{2}{*}{\textbf{Scenes}} & \multicolumn{3}{c|}{\textbf{NeuS \cite{neus}}}     & \multicolumn{3}{c|}{\textbf{Instant-NGP \cite{instant_ngp}}} & \multicolumn{3}{c}{\textbf{NeuS2 \cite{neus2}}}  \\
                   \textbf{}    &  \textbf{}    & PSNR $\uparrow$ & SSIM $\uparrow$ & LPIPS $\downarrow$  & PSNR $\uparrow$  & SSIM $\uparrow$ & LPIPS $\downarrow$  & PSNR $\uparrow$ & SSIM $\uparrow$ & LPIPS $\downarrow$ \\
\midrule
\multirow{8}{14mm}{Dance} & 1080\_Dance\_Dunhuang\_Pair\_f14f15   & 21.38 & 0.959 & 0.042   & 26.37 & 0.974 & 0.035      & 25.34 & 0.967  & 0.044 \\
                          & 1080\_Dance\_Dunhuang\_Single\_f12    & 25.19 & 0.978 & 0.024   & 31.46 & 0.983 & 0.019      & 30.61 & 0.985  & 0.020 \\
                          & 1080\_Dance\_Dunhuang\_Single\_f13    & 25.37 & 0.979 & 0.021   & 33.91 & 0.989 & 0.013      & 31.31 & 0.984  & 0.016 \\
                          & 1080\_Dance\_Dunhuang\_Single\_f14    & 25.03 & 0.977 & 0.025   & 32.02 & 0.988 & 0.016      & 30.34 & 0.985  & 0.016 \\
                          & 1080\_Dance\_Dunhuang\_Single\_f15    & 25.61 & 0.978 & 0.029   & 34.91 & 0.992 & 0.014      & 32.27 & 0.990  & 0.016 \\
                          & 1080\_Dance\_Jazz\_Single\_c22        & 26.59 & 0.984 & 0.018   & 31.41 & 0.987 & 0.011      & 35.38 & 0.991  & 0.010 \\
                          & 1080\_Dance\_Tibetan\_Single\_c22     & 26.26 & 0.984 & 0.020   & 33.69 & 0.991 & 0.016      & 34.28 & 0.990  & 0.014 \\
                          & 1080\_Dance\_Banquet\_Single\_c23     & 26.73 & 0.984 & 0.016   & 36.20 & 0.993 & 0.009       & 35.56 & 0.993  & 0.009 \\   
                          & \textbf{Average}  & 25.27 & 0.978 & 0.024   & \textbf{32.50} & \textbf{0.987} & \textbf{0.017}  & 31.90 & 0.986 & 0.018  \\                                                                                       
\midrule
\multirow{15}{14mm}{Kungfu} & 1080\_Kungfu\_Weapon\_Pair\_m12m13  & 19.51 & 0.941 & 0.061   & 22.31 & 0.955  & 0.014     & 22.07  & 0.962  & 0.048 \\
                            & 1080\_Kungfu\_Double\_Pair\_m12m13  & 18.48 & 0.939 & 0.062   & 20.42 & 0.948  & 0.144     & 22.45  & 0.965  & 0.047 \\
                            & 1080\_Kungfu\_Basic\_Pair\_c24c25   & 25.48 & 0.979 & 0.024   & 31.96 & 0.989  & 0.017     & 30.99  & 0.986  & 0.019 \\
                            & 1080\_Kungfu\_Fan\_Single\_m12      & 25.72 & 0.981 & 0.024   & 33.16 & 0.988  & 0.021     & 30.56  & 0.985  & 0.021 \\
                            & 1080\_Kungfu\_Taichi\_Single\_m12   & 22.16 & 0.976 & 0.027   & 30.94 & 0.988  & 0.020     & 29.60  & 0.986  & 0.020 \\
                            & 1080\_Kungfu\_Shaolin\_Single\_m12  & 23.01 & 0.978 & 0.029   & 31.85 & 0.989  & 0.017     & 32.08  & 0.989  & 0.016 \\
                            & 1080\_Kungfu\_Sword\_Single\_m13    & 23.24 & 0.978 & 0.023   & 31.10 & 0.986  & 0.019     & 29.39  & 0.985  & 0.019 \\
                            & 1080\_Kungfu\_Spear\_Single\_m13    & 22.53 & 0.973 & 0.033   & 31.85 & 0.989  & 0.018     & 28.16  & 0.985  & 0.021  \\
                            & 1080\_Kungfu\_Kick\_Single\_m13     & 20.16 & 0.970 & 0.032   & 29.43 & 0.986  & 0.032     & 28.68  & 0.987  & 0.024 \\
                            & 1080\_Kungfu\_Basic\_Single\_m13    & 23.17 & 0.978 & 0.021   & 28.18 & 0.982  & 0.024     & 28.53  & 0.985  & 0.019 \\
                        & 1080\_Kungfu\_Tongbeiquan\_Single\_m13  & 23.59 & 0.980 & 0.024   & 32.19 & 0.991  & 0.016     & 31.61  & 0.988  & 0.017 \\
                            & 1080\_Kungfu\_Nunchuck\_Single\_m14 & 21.21 & 0.976 & 0.024   & 29.62 & 0.985  & 0.046     & 28.32  & 0.984  & 0.022 \\
                            & 1080\_Kungfu\_Nanquan\_Single\_c24  & 25.81 & 0.983 & 0.026   & 37.62 & 0.995  & 0.013     & 34.67  & 0.993  & 0.014 \\
                        & 1080\_Kungfu\_Broadsword\_Single\_c24   & 25.65 & 0.985 & 0.018   & 35.28 & 0.993  & 0.014     & 37.24  & 0.994  & 0.009 \\
                            & 1080\_Kungfu\_Boxing\_Single\_c25   & 24.31 & 0.981 & 0.024   & 38.37 & 0.996  & 0.009     & 36.91  & 0.995  & 0.009 \\
                            & \textbf{Average} & 22.94 & 0.973 & 0.030  & \textbf{30.95} & 0.984  & 0.028 & 30.08 & \textbf{0.985} & \textbf{0.022}  \\                                                                               
\midrule
\multirow{3}{14mm}{Sport}   & 1080\_Sport\_Football\_Single\_m11  & 24.91 & 0.983 & 0.017   & 29.83 & 0.982 & 0.018      & 30.50   & 0.986  & 0.016 \\
                          & 1080\_Sport\_Taekwondo1\_Pair\_m11c21 & 23.72 & 0.970 & 0.037   & 32.50 & 0.989 & 0.019      & 27.12   & 0.981  & 0.029 \\
                            & 1080\_Sport\_Badminton\_Single\_f11 & 25.22 & 0.980 & 0.028   & 34.29 & 0.993 & 0.011      & 33.79   & 0.993  & 0.014 \\
                            & \textbf{Average} & 24.62 & 0.977 & 0.028   & \textbf{32.20} & \textbf{0.988} & \textbf{0.016} & 30.47 & 0.987 & 0.020  \\                                      
\midrule
\multirow{7}{14mm}{Fashion Show} & 4K\_Studios\_Show\_Pair\_f16f17 & 23.26 & 0.977 & 0.036  & 35.02 & 0.991 & 0.020       & 32.12  & 0.987  & 0.025 \\
                                 & 4K\_Studios\_Show\_Pair\_f18f19 & 22.80 & 0.976 & 0.031  & 34.24 & 0.993 & 0.016       & 32.23  & 0.992  & 0.014 \\
                                  & 4K\_Studios\_Show\_Single\_f16 & 20.38 & 0.975 & 0.042  & 34.49 & 0.990 & 0.012       & 34.33  & 0.993  & 0.012 \\
                                  & 4K\_Studios\_Show\_Single\_f17 & 23.07 & 0.982 & 0.036  & 35.08 & 0.992 & 0.021       & 34.11  & 0.992  & 0.018 \\              
                                  & 4K\_Studios\_Show\_Single\_f18 & 22.95 & 0.983 & 0.027  & 36.73 & 0.995 & 0.012       & 34.32  & 0.994  & 0.013 \\              
                                & 4K\_Studios\_Show\_Single\_f19   & 24.36 & 0.986 & 0.024  & 38.94 & 0.996 & 0.010       & 37.03  & 0.995  & 0.011 \\              
                                 & 4K\_Studios\_Dance\_Single\_f20 & 22.67 & 0.981 & 0.028  & 30.50 & 0.986 & 0.026       & 31.67  & 0.989  & 0.018 \\
                                 & \textbf{Average} & 22.79 & 0.980 & 0.032  & \textbf{35.00} & \textbf{0.992} & 0.017    & 33.69 & \textbf{0.992} & \textbf{0.016}  \\                                      
\midrule
\multicolumn{2}{c|}{\textbf{Average}}        & 23.90 & 0.977 & 0.029  & \textbf{32.66} & \textbf{0.988} & \textbf{0.019}       & 31.53 & 0.987 & \textbf{0.019} \\
\bottomrule
\end{tabular}
\end{table*}

\subsection{Dynamic Human Modeling}
{\bf More Analyses of Dynamic Human Modeling.} Free-viewpoint rendering of a moving subject from a monocular self-rotating video is a complex yet desirable setup. In the 4K Studio sequences category, we provide monocular self-rotating videos of human performers. These videos demonstrate the versatility of our dataset in synthesizing novel views of dynamic humans from fixed monocular cameras. To further illustrate this, we conducted additional experiments using HumanNeRF \cite{mono_humannerf} baseline, a free-viewpoint rendering method for a moving subject. We selected 4 scenes from the 4K Studio sequences category with diverse motions and appearances and used images captured by camera 27, resulting in sequences ranging from 250 to 300.

\begin{figure*}[h]
  \centering
   \includegraphics[width=\linewidth]{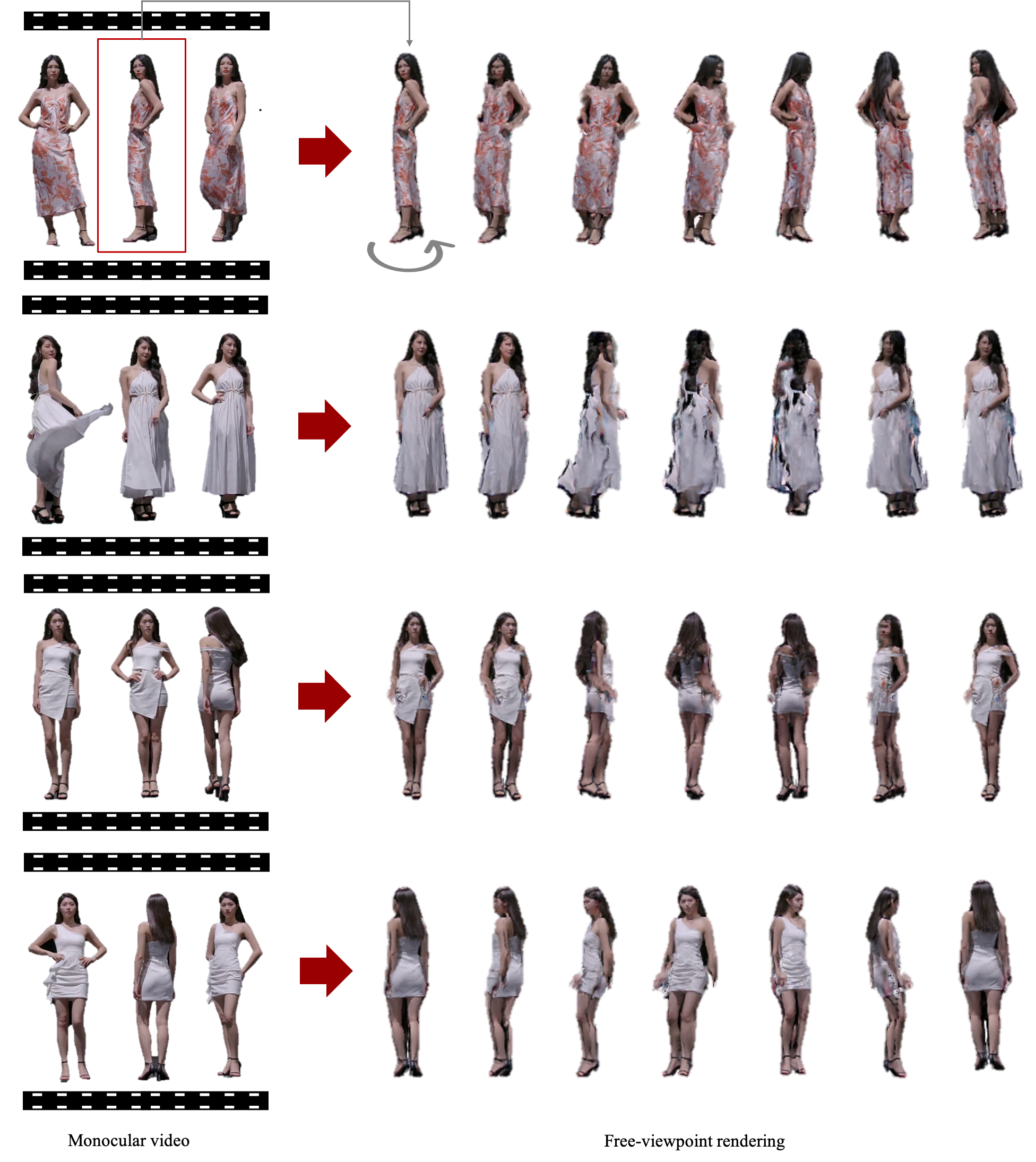}
   \caption{Visual examples of free view synthesis on the four scenes data are provided. The input is a monocular video capturing a human performing complex movements (left). The HumanNeRF \cite{mono_humannerf} generates a free-viewpoint rendering for any frame in the sequence (right).}
   \label{fig:mono_human}
\end{figure*}

We provide four visual examples of our challenging scenario dataset in \cref{fig:mono_human}. While body pose and non-rigid motion were not completely recovered, as the movement of the skirts relied on the temporal dynamics of subject motion. We hope the result points in a promising direction towards modeling humans in complex poses and clothing, and eventually achieving fully photorealistic, freeviewpoint rendering of moving people.

\end{document}